\documentclass[a4paper,fleqn]{cas-dc}
\usepackage[square, numbers, comma, sort&compress]{natbib}
\usepackage{setspace}
\usepackage{listings}
\usepackage{xcolor}
\usepackage[ruled]{algorithm2e}
\usepackage{graphicx}
\usepackage{amsmath}
\usepackage{amssymb}
\usepackage{mathrsfs}
\usepackage{amsthm}
\usepackage{wrapfig, blindtext}
\usepackage[utf8x]{inputenc}
\usepackage[noend]{algpseudocode}
\usepackage{caption}
\usepackage{subcaption}
\usepackage{bm}
\usepackage{float}
\usepackage{nccmath}
\usepackage[acronym]{glossaries}
\usepackage{hyperref}

\makeglossaries
\glsdisablehyper
\newacronym{nn}{NN}{Neural Network}
\newacronym{fcn}{FCN}{Fully Convolutional Network}
\newacronym[plural=CNNs, firstplural = Convolutional Neural Networks (CNNs)]{cnn}{CNN}{Convolutional Neural Network}
\newacronym{ml}{ML}{Machine Learning}
\newacronym{yolo}{YOLO}{You Only Look Once}
\newacronym{nms}{NMS}{Non-Maximum Suppression}
\newacronym{ccps}{CCPs}{Conditional Class Probabilities}
\newacronym{ccp}{CCP}{Conditional Class Probability}
\newacronym{csc}{CSC}{Class Specific Confidence}
\newacronym{iou}{IOU}{Intersection Over Union}
\newacronym{ap}{AP}{Average Precision}
\newacronym{map}{mAP}{mean Average Precision}
\newacronym{ai}{AI}{Artificial Intelligence}
\newacronym{pca}{PCA}{Principal Component Analysis}

\def\tsc#1{\csdef{#1}{\textsc{\lowercase{#1}}\xspace}}
\tsc{WGM}
\tsc{QE}
\tsc{EP}
\tsc{PMS}
\tsc{BEC}
\tsc{DE}

\begin{document}
\let\WriteBookmarks\relax
\def\floatpagepagefraction{1}
\def\textpagefraction{.001}
\shorttitle{Object Detection on Marine Data}
\shortauthors{H.  Stavelin et~al.}
\title [mode = title]{Marine life through You Only Look Once's perspective} 
\author[1]{Herman Stavelin}
\ead{hermanstavelin@gmail.com}
\author[1,2]{Adil Rasheed}
\cormark[1]
\ead{adil.rasheed@ntnu.no}
\ead[url]{www.adilrasheed.com}
\author[3]{Omer San}
\ead{osan@okstate.edu}
\author[4]{Arne Johan Hestnes}
\ead{arne.hestnes@km.kongsberg.com}

\address[1]{Norweigan University of Science and Technology, Elektro D/B2, 235, Gløshaugen, O. S. Bragstads plass 2, Trondheim, Norway}
\address[2]{Mathematics and Cybernetics, SINTEF Digital, Klæbuveien 153, Trondheim, Norway}
\address[3]{Oklahoma State University, 201 General Academic Building
Stillwater, Oklahoma 74078 USA}
\address[4]{Kongsberg Maritime, Strandpromenaden 50, Horten, Norway}

\begin{abstract}
With the rise of focus on man made changes to our planet and wildlife therein, more and more emphasis is put on sustainable and responsible gathering of resources. In an effort to preserve maritime wildlife the Norwegian government has decided that it is necessary to create an overview over the presence and abundance of various species of wildlife in the Norwegian fjords and oceans. In this paper we apply and analyze an object detection scheme that detects fish in camera images. The data is sampled from a submerged data station at Fulehuk in Norway. We implement You Only Look Once (YOLO) version 3 and create a dataset consisting of 99,961 images with a mAP of $\sim 0.88$. We also investigate intermediate results within YOLO, gaining insight into how it performs object detection.
\end{abstract}
\maketitle


\section{Introduction}
Every year more and more maritime life disappears from the Norwegian fjords and oceans. In order to combat this development, the Norwegian government has launched a project called Frisk Oslofjord - Healthy Oslo Fjord \cite{OmOSLOFJORD}. One of the main goals of the project is to prepare detailed ecological maps of the Oslo fjord. These maps are expected to show the class of marine species and their locations at any particular time. As of today this procedure is conducted manually by inspecting images and then recording the findings. However, with the recent success of \gls{ai} and \gls{ml} in image classification, text interpretation and big data analysis, new possibilities are opening up to address relevant questions. For example, in \cite{olsvik2019biometric}, \cite{snumed}, and \cite{waterpower_2018}, authors have already shown the power of computer vision and \gls{ml}, not only in identifying, but also classifying various marine species. The approach, owing to the ease of automation, will allow mapping of the fjords and ocean in general, with much higher spatio-temporal resolutions. However, despite the huge potential of exploitation of the \gls{ml} based approach, the technology is not perfect \cite{disadvantagesNN} \cite{abbe2018provable}. We refer to \cite{moniruzzaman2017deep} for a survey of deep learning methods on underwater marine object detection and automated approaches for monitoring of underwater ecosystem including seagrass meadows. The algorithms which can give super-human performance in image classification in good daylight might suffer to make correct classifications in underwater scenarios where the visibility is highly diminished due to poor lighting conditions. 

Recently, a tiered observation system with improved imaging capabilities has been promoted for tracking the status and trends in marine macrophyte cover \cite{duffy2019toward}. For example, detection of coral reef fishes in underwater images has been studied using convolutional neural networks \cite{villon2016coral,villon2018deep}, where different post-processing decision rules to identify 20 fish species have been taken into account for monitoring marine biodiversity in a cost-effective manner. In their analyses built-up new video-based protocols, the authors highlighted the promise of deep learning for monitoring fish biodiversity cheaply and effectively with an identification accuracy of 94.9\%, which is greater than the rate of correct identification by humans (89.3\%). A fish detection system under a variety of benthic background and illumination conditions has been introduced using a cascaded deep networks by combining convolutional nets and long short-term memory networks \cite{labao2019cascaded}. Based on a real-world fish recognition dataset equipped by a linear support vector machine classifier, \citeauthor{sun2018low}\cite{sun2018low} also introduced a two-stage principal component analysis network to extract features from fish images. 

The current work in this regard attempts to answer some important questions related to the application of \gls{ml} to real life problems like the one undertaken here. In particular we will demonstrate a semi-supervised way of labeling data, the application of an algorithm to detect fish in images collected under noisy conditions, and then finally giving an insight into the inner workings of the algorithm. To this end, we start this paper with a brief overview of the theory behind the algorithm we used for object detection followed by information regarding the collection and processing of data. After this, results related to the prediction capability of the \gls{ml} algorithm are presented followed by some insight into its inner workings. Lastly we present the main take away from the current study and propose future course of research. 

\section{Theory} 
\subsection{YOLO}
The \gls{ml} algorithm which we have utilized is based on the \gls{yolo} algorithm which is one of the most efficient and accurate algorithm for object detection in complicated scenes \cite{redmon2018yolov3}. So far, the algorithm has been adopted in many applications including chemical sensing and detection of gas emission \cite{monroy2018semantic}, anthracnose lesion detection on plant surfaces \cite{tian2019detection}, small target detection from drones \cite{xu2018vehicle}, traffic monitoring \cite{barthelemy2019edge}, plate recognition \cite{laroca2018robust}, pedestrian detection \cite{qu2018pedestrian}, and autonomous driving applications \cite{choi2019gaussian}.  Since, one of the objectives of the current study is to present an insight into the inner workings of the algorithm we give a brief, but sufficient description of the algorithm. \gls{yolo} is a \gls{fcn} \cite{redmon2018yolov3}. It uses a feature extractor with residual blocks consisting of 53 convolutional layers. One unique feature of this algorithm is that the detections are 
conducted at different depths through the network.

\begin{figure*}
    \includegraphics[width=\textwidth]{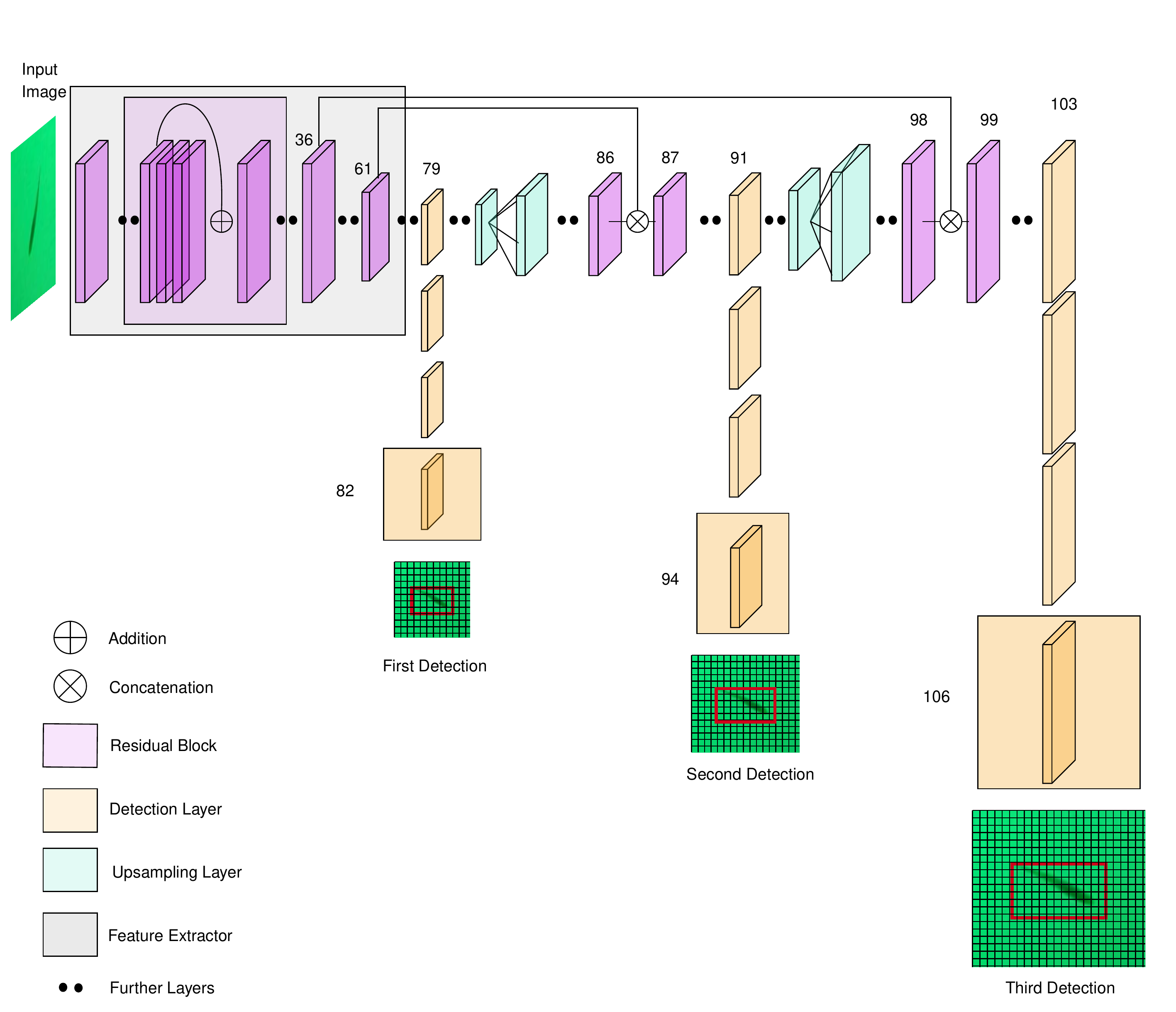}
    \caption[The structure of the entire YOLO v3 network]{The structure of the entire YOLO v3 network. Illustration inspired by \cite{tds_yolov3_webarticle}}
    \label{fig:yolov3_architecture}
\end{figure*}
In Figure \ref{fig:yolov3_architecture} the entire structure of the network is shown. On the far left of the network one can see the layer through which the input images are fed in. This is followed by a gray box indicating YOLO's feature extractor. The feature extractor, as the name implies, is responsible for extracting features from the input. It consists of 23 residual blocks, each of which are built up of convolutional layers with $3 \times 3$ and $1 \times 1$ kernels. Batch normalization is applied in every convolutional layer to regularize the model, thus avoiding overfitting without the invocation of dropout (\cite{redmon2016yolo9000}). $3 \times 3$ kernels with stride 2 are used when downsampling the feature map. \gls{yolo} uses no form of pooling in contrast to most other FCNs \cite{zhao2018object}. This is because pooling is often attributed to loss of low-level features \cite{yoloPytorch_webarticle}. 

\begin{figure}
    \centering
    \includegraphics[width=0.9\columnwidth]{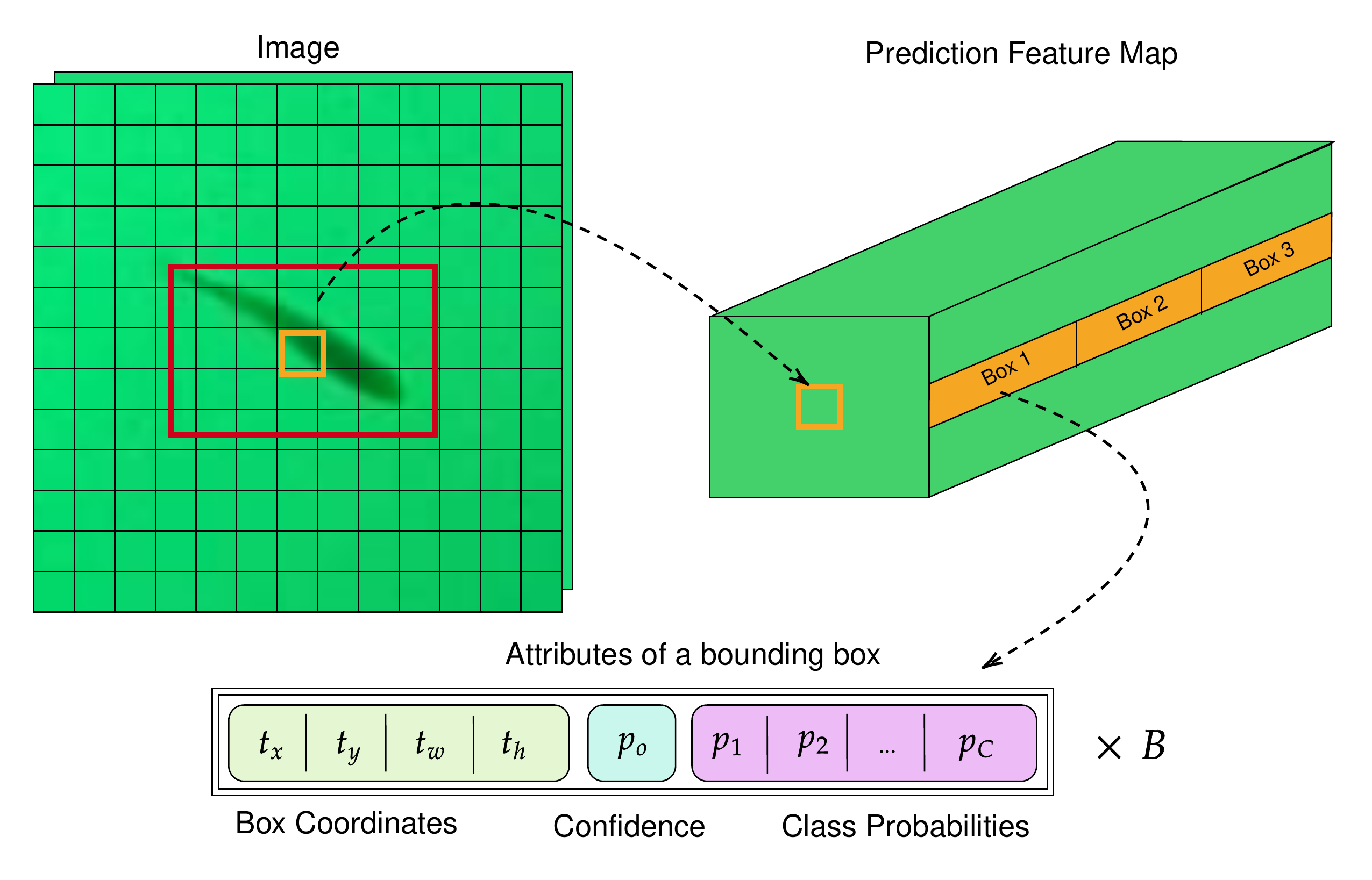}
    \caption[Explanation of YOLOs output tensor]{Explanation of YOLOs output tensor. Illustration inspired by \cite{yoloPytorch_webarticle}}
    \label{fig:dog}
\end{figure}

Since \gls{yolo} is a \gls{fcn}, it is invariant to the size of the input images. However, for mere convenience (for example in batch processing of images and parallelization on GPUs) we have kept the dimensions of all the images the same. Detections are made at layer $82$, layer $94$ and layer $106$. By the time the input image transverse down to the first detection layer, its size shrinks by a factor of $32$. Thus with an input image of size $416 \times 416$ the feature map at this layer will be $13 \times 13$. After the first detection, the layer prior to the detection is upsampled by a factor of $2$. In the figure this corresponds to taking the last purple layer before the first orange layer. After a few more convolutional layers the current layer is concatenated with a feature map from an earlier layer having identical size. In Figure \ref{fig:yolov3_architecture} this is shown as concatenation and we see that layer 61 and 86 are concatenated to produce layer 87. Then, at layer 94, YOLO again extracts detections. The exact same procedure repeats once more. If the input image was $416 \times 416$, the feature maps in layer $94$ and $106$ would be of size $26$ and $52$ respectively. Extraction of detections at three locations is an added feature of the third version of YOLO. According to the authors it improves the detection of small objects since it is able to capture more fine-grained features \cite{redmon2018yolov3}. The output of the network is formulated as a 3D tensor and its dimensions are presented in \autoref{eq:yolo_output}. 
\begin{equation}
    \label{eq:yolo_output}
    \text{Output} = S \times S \times [B * (5 + C)]
\end{equation}
where $S$ is the number of grid-cells, $B$ the bounding boxes per grid cell and $C$ the number of classes to detect. In \autoref{fig:dog} we see an illustration of a feature map in a detection layer. A bounding box is displayed as a red rectangle and the orange square is the grid cell that is at the center of the bounding box. This cell contains a long row of values.  $(t_x, t_y)$ are the center of the box relative to the bounds of the grid cell the box belongs to. $(t_w, t_h)$ are the width and height of the box relative to the whole image. The confidence score $p_o$, sometimes called objectness score, tells us how certain it is that there is an object inside a box and also how accurately the box encloses the object. Formally we have: 
\begin{equation}
    \label{eq:conf}
    p_o =  Pr(Object) * IOU_{pred}^{truth}.
\end{equation}
\gls{iou} is mathematically defined in \autoref{eq:iou}. If there is no object inside the box the confidence should be zero. If there is an object in the cell the confidence score should equal the \gls{iou}. The class probabilities $p_i$ are formally defined in \autoref{eq:ccp}.
\begin{equation}
    \label{eq:ccp}
    p_i = Pr(Class_i | Object)
\end{equation}
These probabilities are called \gls{ccp} and are a measure of how likely it is, given that there is an object, that this object belongs to a certain class. Prior to the third version of \gls{yolo} softmax was applied on the output of the conditional class probabilities \cite{redmon2015look}. This was changed to the sigmoid activation function. This is because the sigmoid allows for objects to belong to several classes. Thus, if one has the classes cod and animal, a fish can belong to both of these classes \cite{redmon2018yolov3}.

\gls{yolo} produces an output tensor like this for each of the three detection layers in the network. The final result is produced by simply adding the outputs together.

\subsection{Evaluation metrics}
There are many metrics that can be used when evaluating the quality of an \gls{ml} algorithm. Here, we will present the most common, ubiquitous metrics. The definitions are obtained from \cite{metricBook2013}, \cite{evaluationML2019metric} and \cite{Everingham10}.

\begin{figure}
    \centering
    \includegraphics[width=1\columnwidth]{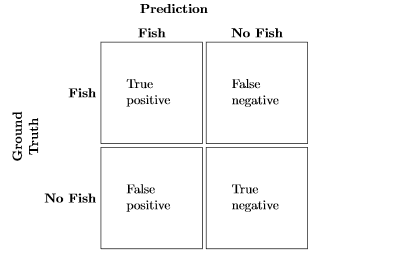}
    \caption{A confusion matrix relating TP, TN, FP and FN}
    \label{tikz:metrics}
\end{figure}

In order to get a pleasing understanding of what these metrics are trying to convey, we will construct a simple scenario. Lets say we have an algorithm that predicts the location and class of some fish within an image. We have some labeled images verifying whether a region of an image actually is a fish or not and its position. If this region contains a fish and our algorithm predicts that this is the case, with some satisfying degree of correctness, we have a so called True Positive (TP). If we predict that there is no fish in the region and our labeled data agrees with this we have a True Negative (TN). False Positives (FP) tells us that we have predicted the presence of a fish when there is none. False Negatives (FN) tells us that we have predicted the absence of a fish when there actually is a fish there. In \autoref{tikz:metrics} we have made a table displaying all these possibilities.

Accuracy, given by \autoref{eqn:accuracy}, is the ratio of correct predictions to all the predictions. 
\begin{equation}
    \text{Accuracy} = \frac{TP + TN}{TP + TN + FP + FN}
    \label{eqn:accuracy}
\end{equation}
Precision, given by \autoref{eqn:precision}, is a measurement of how precise the predictions are. It gives us the percentage of predictions that agree with the ground truth. 
\begin{equation}
    \text{Precision} = \frac{TP}{TP + FP}
    \label{eqn:precision}
\end{equation}

Recall, given by \autoref{eqn:recall}, tells us how good the algorithm is at finding all the TPs in an image. 
\begin{equation}
    \text{Recall} = \frac{TP}{TP + FN}
    \label{eqn:recall}
\end{equation}
Precision and recall are strongly related. High precision and low recall means that we are very sure that the objects we have detected actually are correct. Low precision and high recall means that we have found all the objects in the image but we also labeled a lot of junk as fish. For most applications we want to find the parameters that lead to the best combined precision and recall. The F1-score (\autoref{eqn:F1-score}) is designed for exactly this - a combination of precision and recall. 
\begin{equation}
    \text{F1-score} = 2\cdot\frac{\text{Precision} \cdot \text{Recall}}{\text{Precision} + \text{Recall}}
    \label{eqn:F1-score}
\end{equation}
In many applications we want to optimize both with respect to precision and to recall. Thus it is often simpler to just treat the F1-score.
\begin{equation}
    \text{IOU} = \frac{\text{area of overlap}}{\text{area of union}}
    \label{eq:iou}
\end{equation}
\gls{iou} (\autoref{eq:iou}) is a metric for how much two shapes overlap. A high \gls{iou} means that the two shapes almost perfectly overlap each other. An \gls{iou} of zero would correspond to the two shapes not overlapping at all.
\begin{equation}
    \text{AP@IOU} = \sum_{n}\frac{\text{Recall}_n - \text{Recall}_{n-1}}{\text{Precision}_n}
    \label{eq:ap@iou}
\end{equation}

\begin{figure}
    \centering
    \includegraphics[width=1\columnwidth]{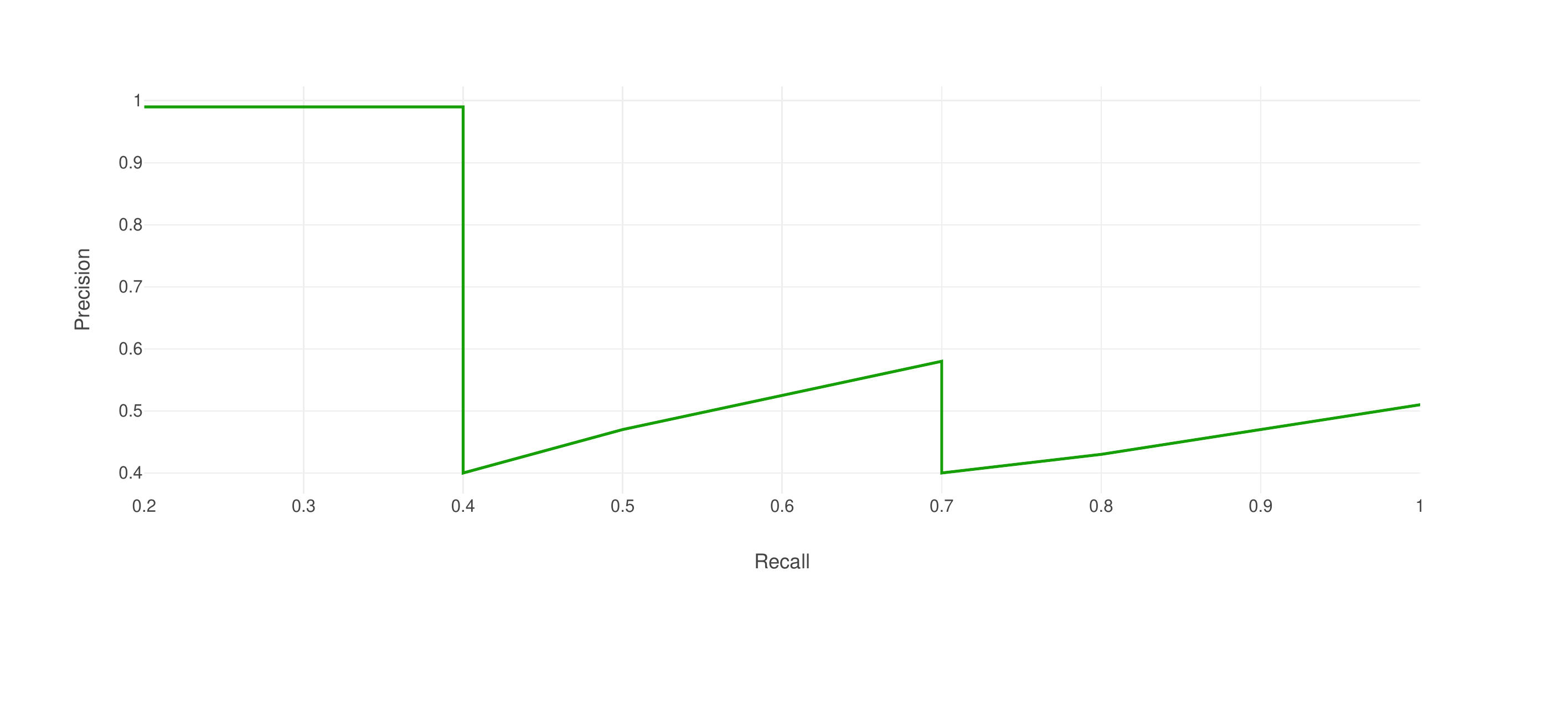}
    \caption[Precision-recall curve]{Precision-recall curve. Illustration inspired by \cite{precision_recall_map}}
    \label{fig:precision-recall}
\end{figure}

There are many different formulations of \gls{ap}. We have chosen to formulate it the same way that scikit-learn does \cite{scikit-learn}. The @IOU refers to the fact that one must define when a prediction is accurate enough that its a TP. For example how precisely a bounding box must encapsulate an object for there to be detection. This is done by setting an \gls{iou} threshold. Typically the \gls{ap} is calculated for an \gls{iou} $\geq 0.5$. From \autoref{eq:ap@iou} is can be seen that calculating the \gls{ap} is the same as calculating the area under the precision-recall curve. A typical plot of such a precision-recall curve is shown in \autoref{fig:precision-recall}. This curve is made by sorting predictions in descending order of confidence. For every subset of predictions, precision and recall are calculated. 

\begin{equation}
    \text{mAP} = \frac{\sum_{q=1}^{C} AP}{C} \label{eq:mAP}
\end{equation}
\gls{map} is very much the same as the AP@IOU, but \textit{mean} indicates that we calculate the mean average precision across all the classes in our data. For example an algorithm might be really good at detection all the bass in an image but perform poorly when detecting salmon. Thus it will have a high AP for bass and low AP for salmon. By taking the \gls{map} we will get the overall mean for how good the algorithm is at identifying all the classes. \gls{map} must also be calculated given some threshold \gls{iou} but we have left it out to not clutter the equations too much.

\subsection{Principal Component Analysis}

\gls{pca} is an unsupervised technique for identifying patterns in data, and expressing the data in such a way as to to highlight their similarities and differences \cite{pca_tutorial}. Most commonly it is used as a dimensionality reduction method \cite{pca_reviewandrecentdevelopments}. The fundamental idea is to represent a dataset using fewer variables than the original dataset, while retaining as much information as possible. In this approach, eigenvectors of the covariance matrix have the largest variance and called principal components. In practice, these eigenvectors are ordered by the amplitude of corresponding eigenvalues containing main characteristics of the dateset. \gls{pca} is a simple four step procedure.

\begin{enumerate}
    \item Subtract the mean from the dataset
    \item Calculate the covariance matrix
    \item Calculate the eigenvectors and eigenvalues of the covariance matrix
    \item Choose components and form a feature vector
\end{enumerate}

In \autoref{subsec:insight_pca} we apply \gls{pca} to a set of images. There are several ways to do this. We will here give a precise description of how this is completed. We have two goals we want to achieve:

\begin{enumerate}
    \item To visualize as much of the information in the images as possible, without displaying every single one
    \item To find out if the images contain truly distinct data, making them necessary for the network
\end{enumerate}

We treat an image as a variable and the image width and height as samples. Images are naturally two dimensional and thus can not directly be thought of as samples. Thus we unpack this data into a single dimension. We do this by taking one row at a time from an image and appending it to the next row. This is illustrated in \autoref{eq:unpacking}

\begin{align}
    \label{eq:unpacking}
    I &= \begin{bmatrix}
    c_{00} & c_{01} \\
    c_{10} & c_{11} 
    \end{bmatrix}\\
    &= \begin{bmatrix}
    c_{00} & c_{01}  & c_{10} & c_{11}  
    \end{bmatrix}
\end{align}

If we had 32 images of size $13 \times 13$ we group this together such that we have a matrix of dimensions $32 \times (13*13) = 32 \times 169$. This matrix we can apply \gls{pca} to. The result of this is a list of components that contain the variance in the original dataset.  Two main results are worth noting: The first is when one component explains all the variance in the dataset. In this case the images contain a clear pattern. For example that they are all the same image. The second case is when all the components explain similar levels of variance. In this case there is no clear pattern in the images, indicating that they are all very different.

\section{Data}
\begin{figure}[]
  \includegraphics[width=1\columnwidth]{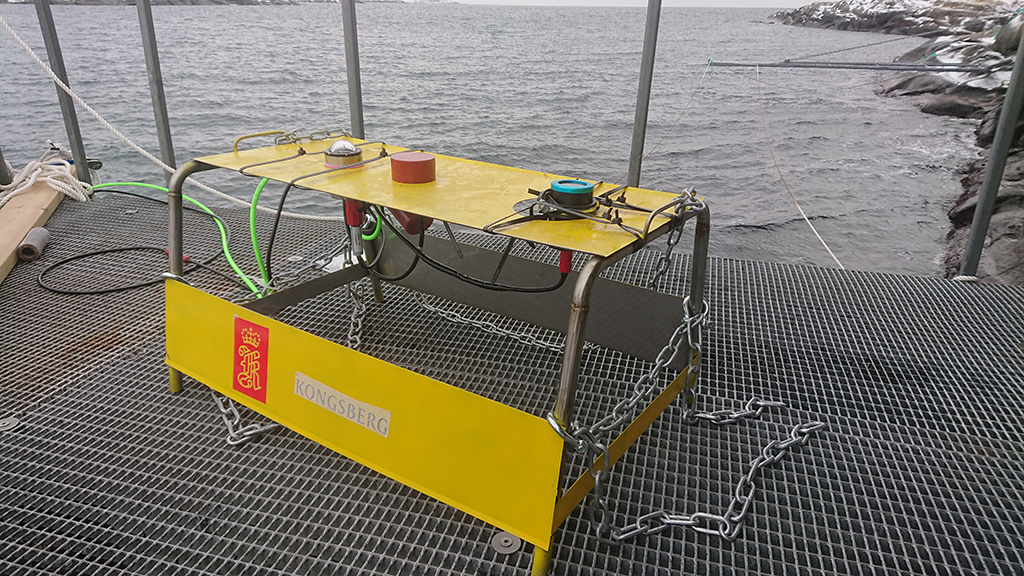}
  \caption[Measurement Station]{The measurement station. The leftmost glass dome is the camera. The red cylinder is the sonar and the blue and black cylinder on the right is the artificial lighting source}
  \label{fig:rigg}
\end{figure}

The measurement station at Fulehuk in Norway can be seen in Figure \ref{fig:rigg}. The station has a camera, a sonar and an artificial lighting source. It was deployed on the ocean floor about $30$ meters below the water surface. It is oriented such that it looks from the ocean floor up at the water surface. The camera is a Goblin Shark and records $1080p$ at $30$ fps with a horizontal angle of view of $92^\circ$ in water \cite{goblin_shark}. The sonar is a Simrad ES200-7CDK Split \cite{sonar}. The data was recorded between March and August 2019. The hardware was initially setup such that the camera and sonar would continuously capture data while artificial lighting would be enabled during nighttime. In order to lessen the data burden, images were uploaded to the storage container at 6 seconds intervals during March and much more infrequently (minutes to hours) during June, July and August. Furthermore, during March there was no artificial lighting during the night. For June, July and August artificial lighting was enabled both during the day and night time. A team of divers occasionally cleaned the camera lens from excessive algae growth.

\section{Implementation and set-up}

\paragraph{}{}
\begin{table}
    \centering
    \begin{tabular}{ll}
    \textbf{Parameters}    & \textbf{Values}         \\ 
    \hline
    batch         & 64          \\
    subdivisions  & 16          \\
    width         & 416         \\
    height        & 416         \\
    channels      & 3           \\
    momentum      & 0.9            \\
    decay         & 0.0005         \\
    angle         & 0         \\
    saturation    & 1.5       \\
    exposure      & 1.5        \\
    hue           & .1             \\
    learning rate & 0.001         \\
    burn\_in      & 1000           \\
    max\_batches  & 500200         \\
    policy        & steps          \\
    steps         & 400000, 450000 \\
    scales        & .1,.1         
    \end{tabular}
    \caption{YOLO parameters}
    \label{table:yolo_parameters}
\end{table}

\begin{figure}
    \centering
    \includegraphics[width=1\columnwidth]{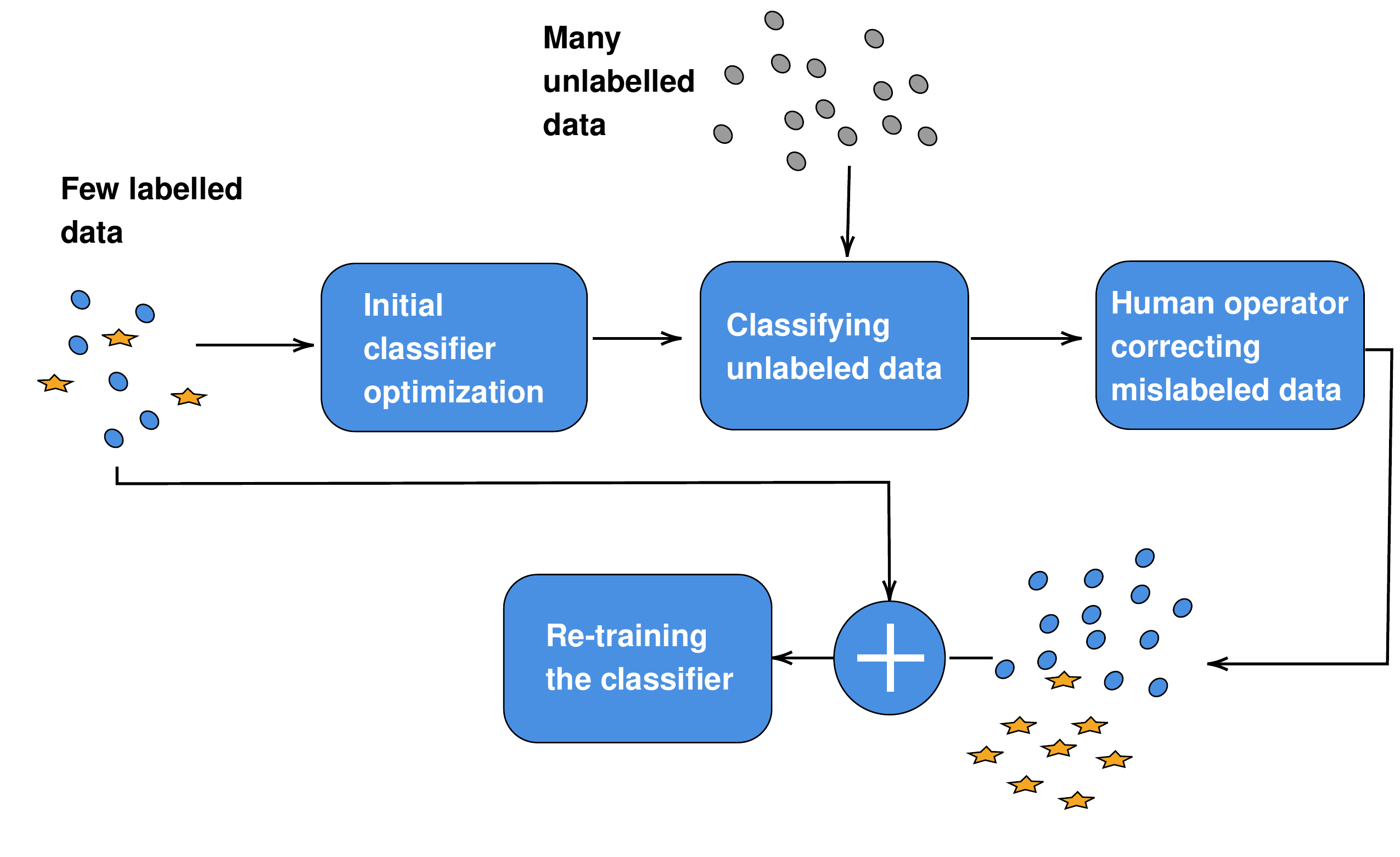} 
    \caption{The training process}
    \label{fig:loop}
\end{figure}

Setting up YOLO requires an initial set of weights and a configuration file. A pretrained set of weights called ``darknet53.conv.74'' were obtained from Redmons website \cite{RedmonYOLOWebsite}. These weights were trained on the ImageNet dataset.  The configuration file contains the entire network structure. All the general parameters are listed in \autoref{table:yolo_parameters}. Width and height were set to $416$ to keep training time low. The number of filters in the three output layers had to be tuned to correspond to the number of classes we wanted to classify. Based on the article \cite{redmon2018yolov3}, the number of filters in the output layers should comply with the following equation:
\begin{equation}
    \text{Filters}_n = (C + 5) * 3
\end{equation}
We started by manually labeling 510 images from the video recorded in March. The data was split in a 90-10 ratio between training and test data. This produced 459 images for training and 51 images for testing. LabelImg was used to to label the images \cite{labelimg}, as it supports the YOLO data format. Only fish class was used in this study. The YOLO algorithm was trained twice. This was done as illustrated in \autoref{fig:loop}. First, the model was trained on a small set of hand labelled images from March. The results from the training are shown in \autoref{table:yolo_results}. The trained model from this stage was then used to pseudo-label 2500 new images from March. Images on which the detections that had a confidence score $\geq 0.25$ were retained while the images corresponding to false positives and false negatives were manually corrected and the network was retrained. Before retraining the network it was initialized with the weights that were learned during the first stage of the training. The results from the second stage of training are shown in \autoref{table:yolo_results_more}.
\section{Results and Discussions}
\begin{figure*}
\centering
    \begin{subfigure}{.45\textwidth}
        \centering
        \includegraphics[width=1\textwidth]{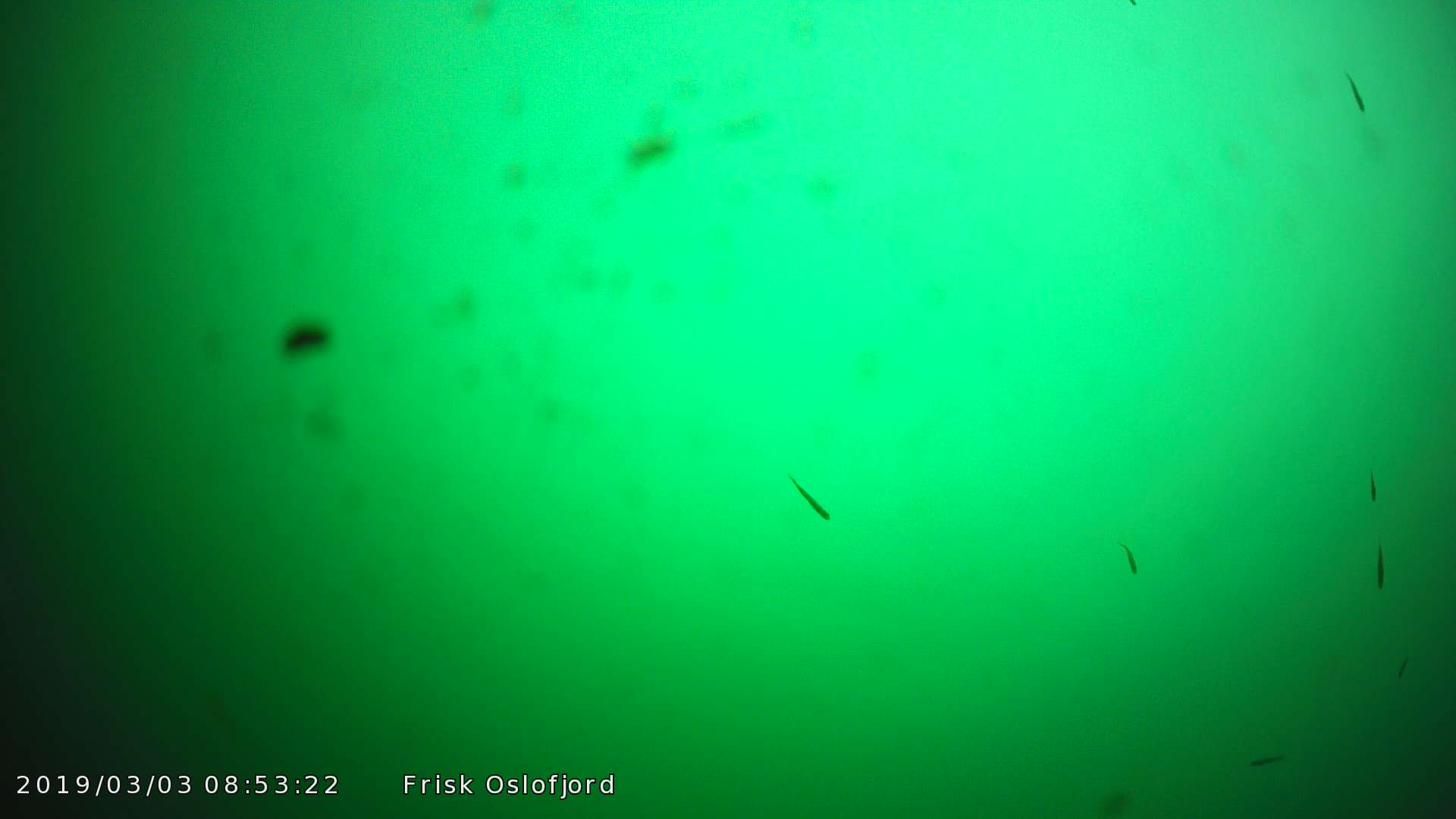}
        \caption{Daytime during March}
    \end{subfigure}%
    \begin{subfigure}{.45\textwidth}
        \centering
        \includegraphics[width=1\textwidth]{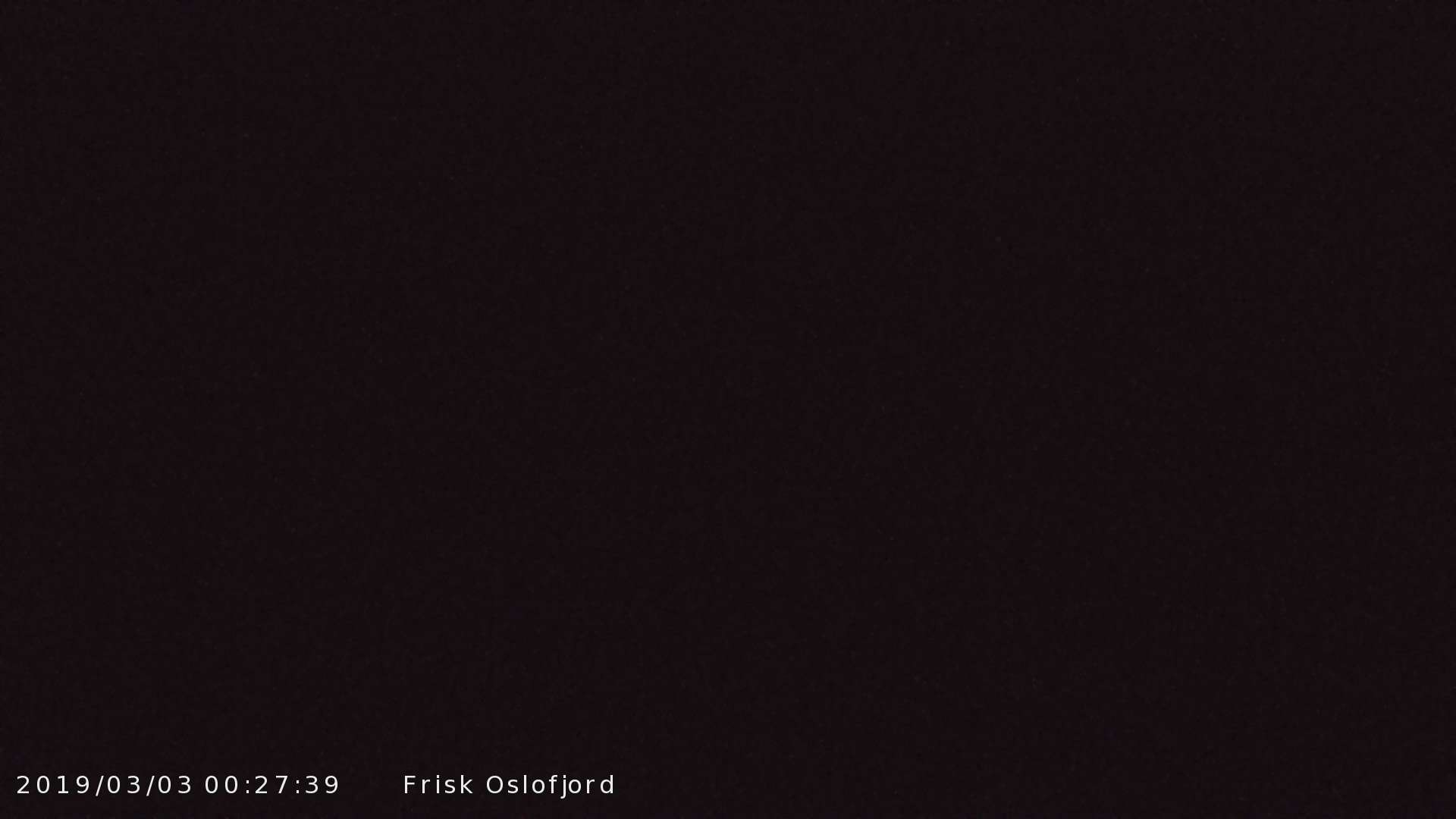}
        \caption{Nighttime during March}
    \end{subfigure}
    \begin{subfigure}{.45\textwidth}
        \centering
        \includegraphics[width=1\textwidth]{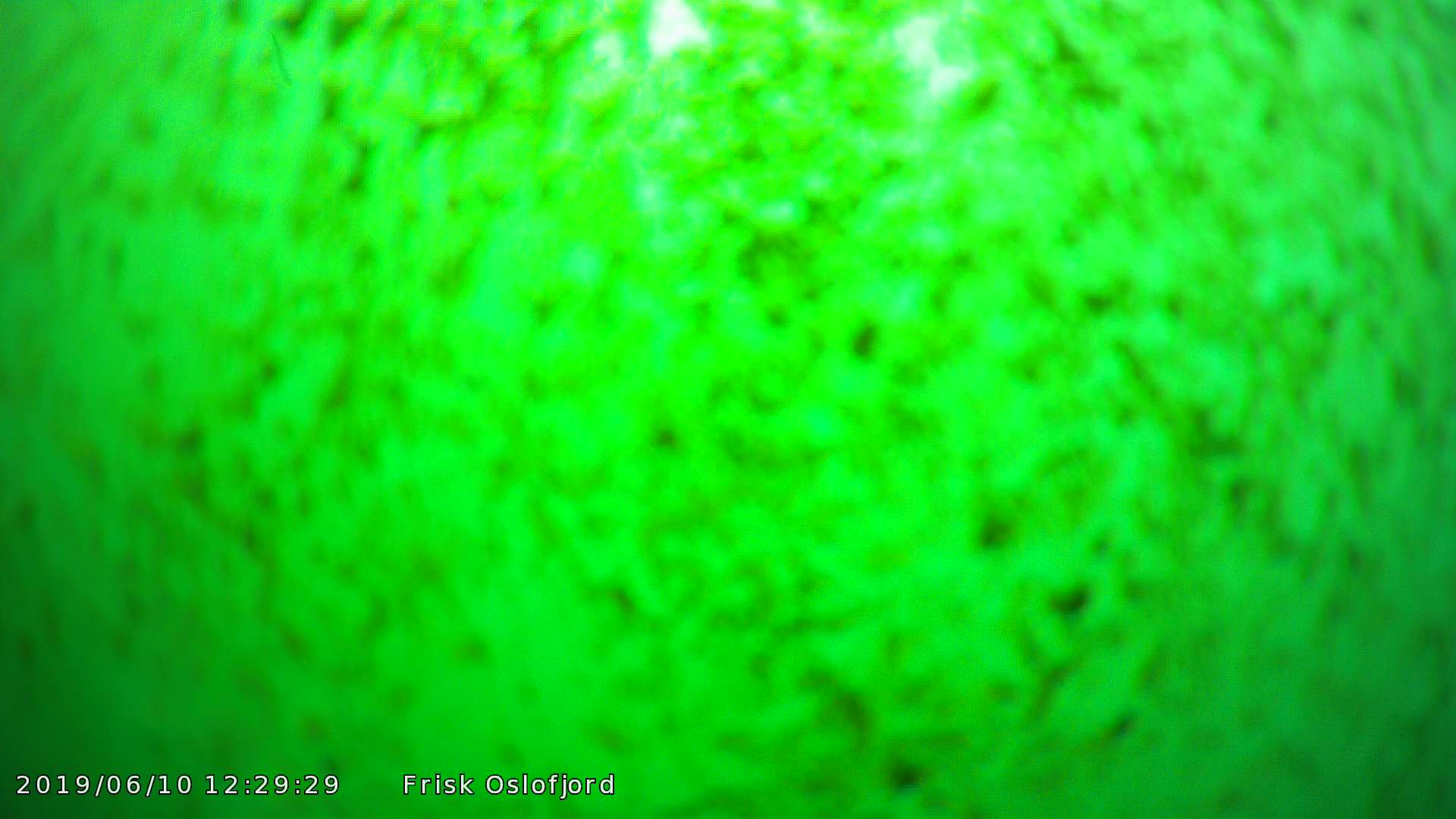}
         \caption{Daytime during June}       
    \end{subfigure}%
    \begin{subfigure}{.45\textwidth}
        \centering
        \includegraphics[width=1\textwidth]{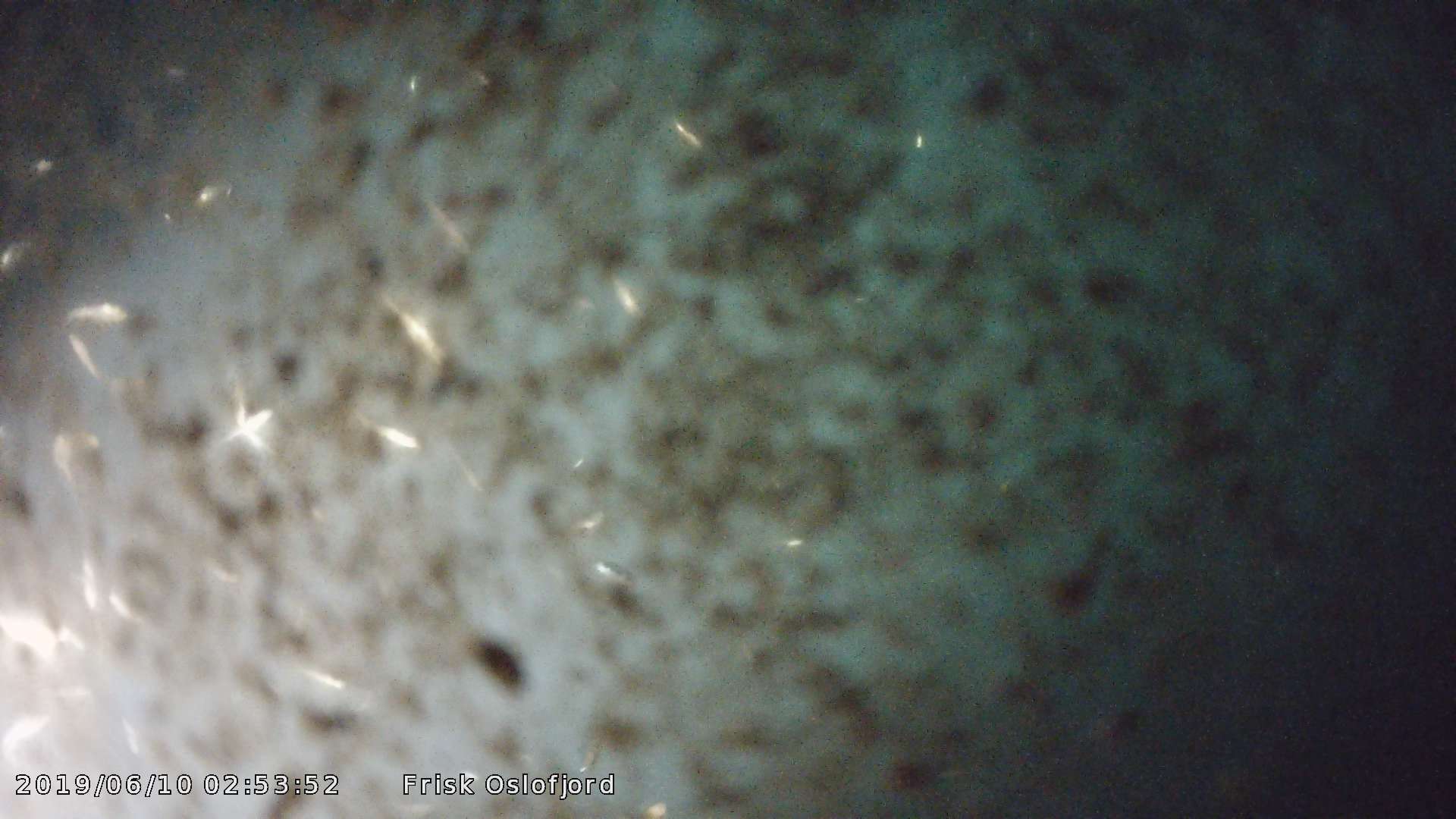}
        \caption{Nighttime during June}
    \end{subfigure}
    \begin{subfigure}{.45\textwidth}
        \centering
        \includegraphics[width=1\textwidth]{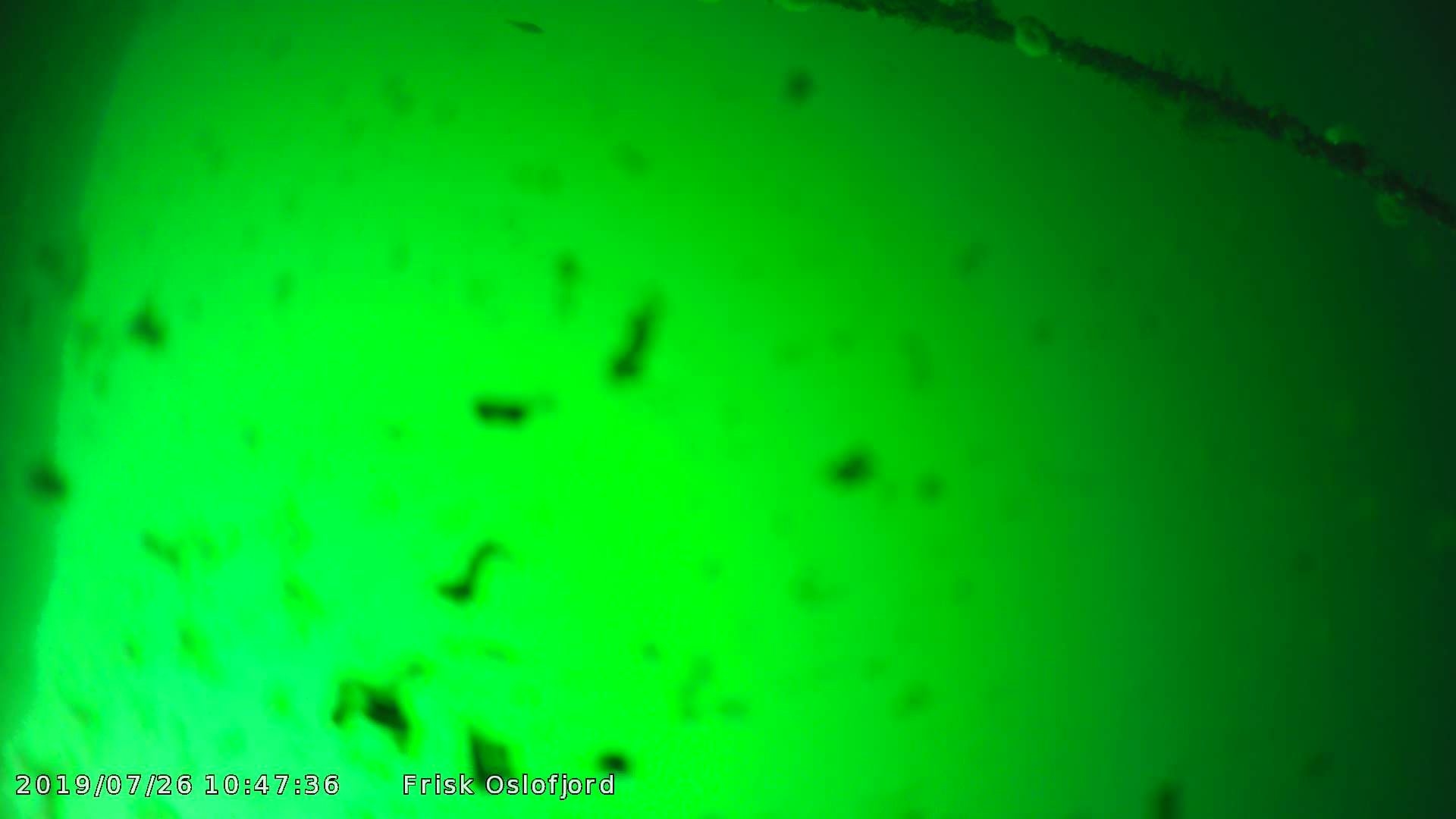}
        \caption{Daytime during July}
    \end{subfigure}%
    \begin{subfigure}{.45\textwidth}
        \centering
        \includegraphics[width=1\textwidth]{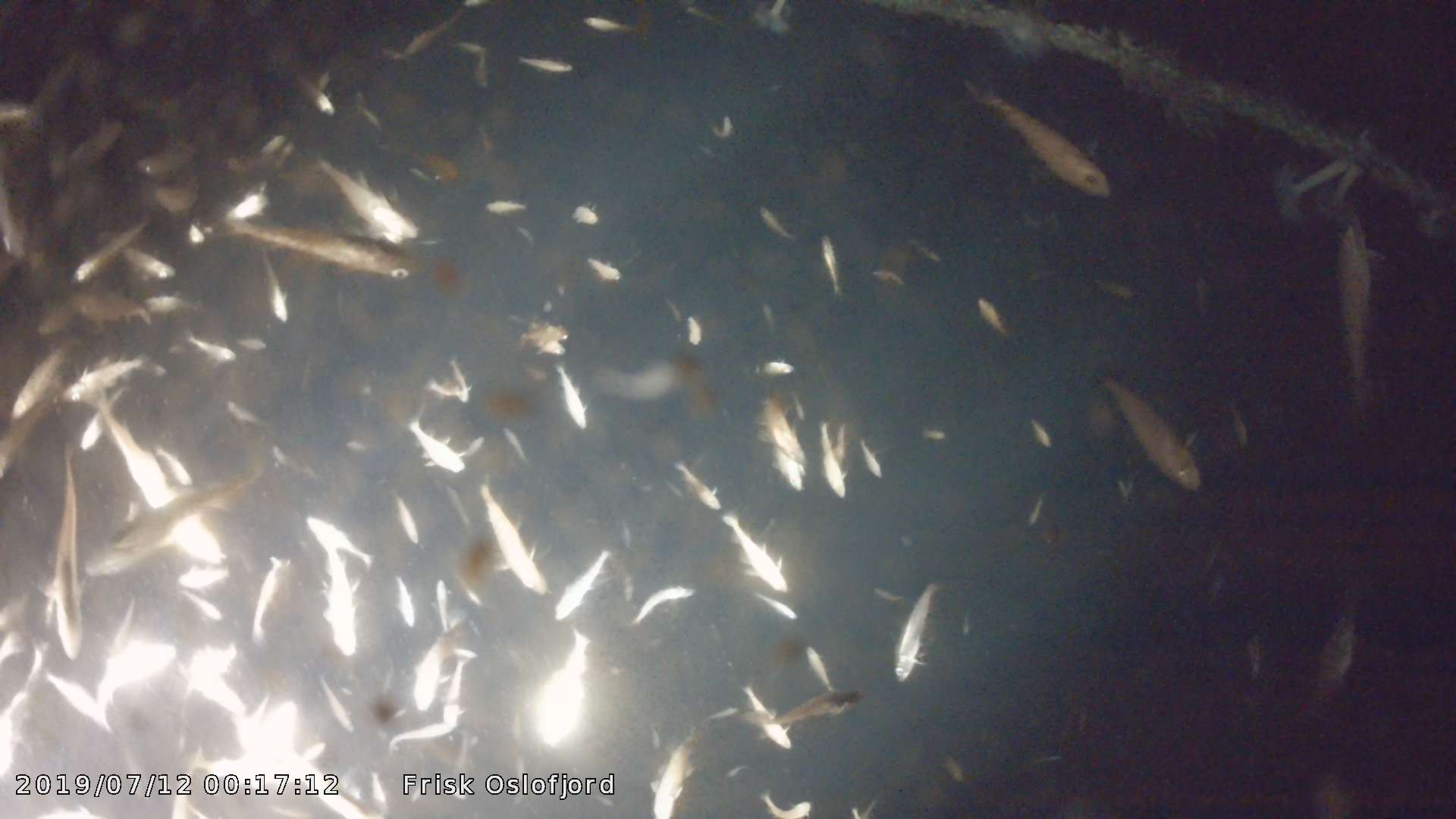}
        \caption{Nighttime during July}
    \end{subfigure}
    \begin{subfigure}{.45\textwidth}
        \centering
        \includegraphics[width=1\textwidth]{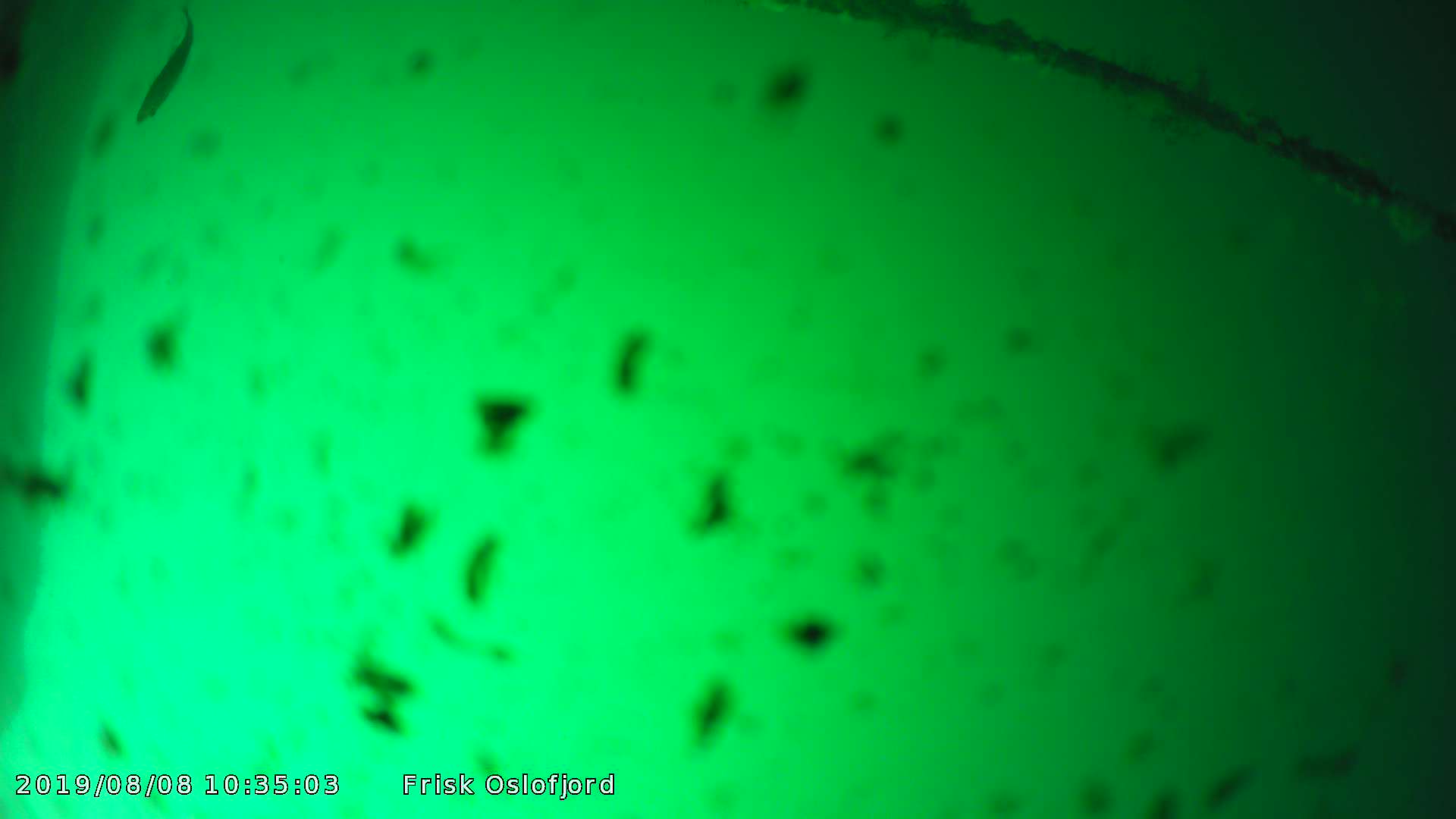}
        \caption{Daytime during August}
    \end{subfigure}%
    \begin{subfigure}{.45\textwidth}
        \centering
        \includegraphics[width=1\textwidth]{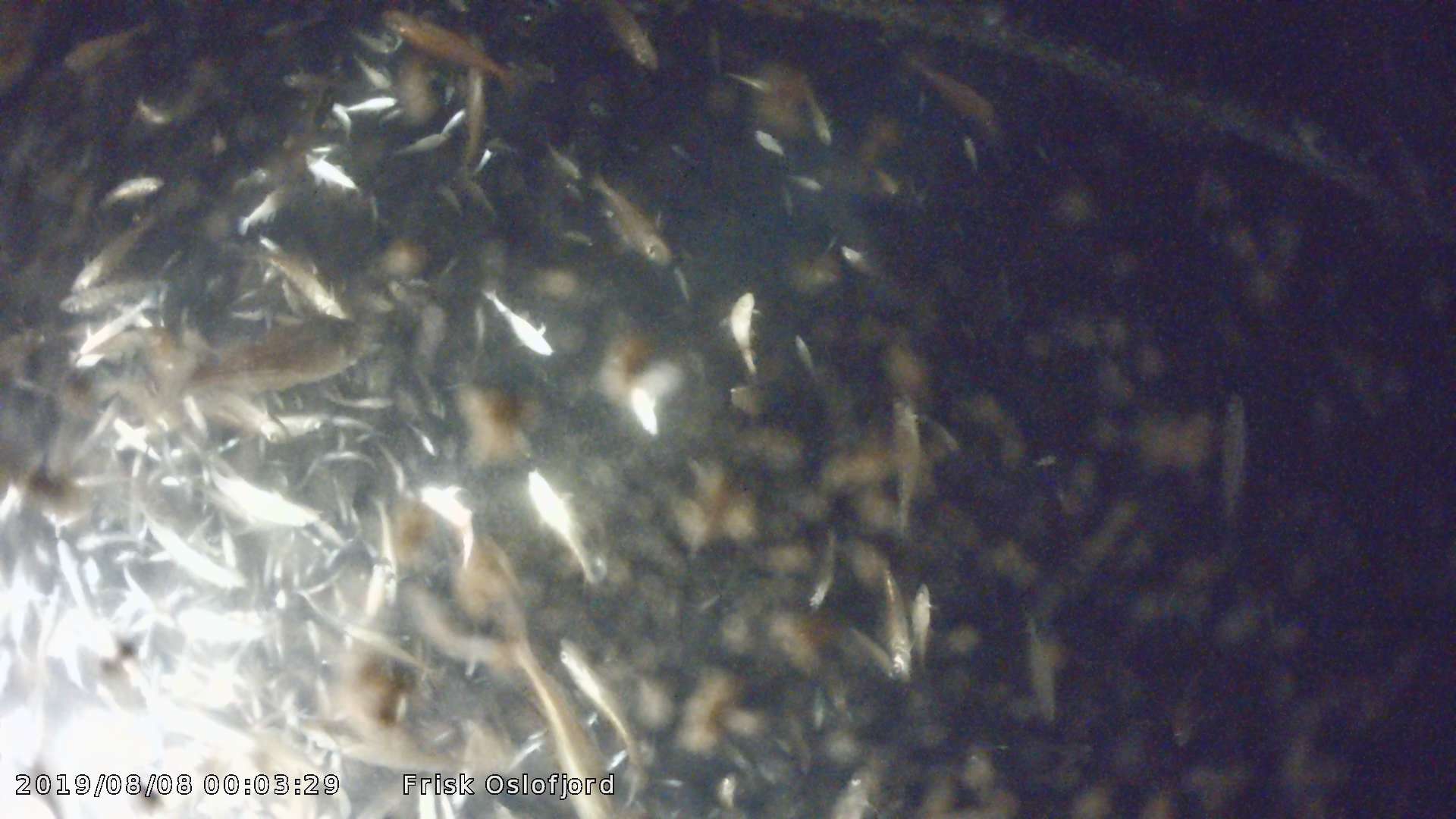}
        \caption{Nighttime during August}
        \label{fig:aug_night_no_det}
    \end{subfigure}
    \caption[Images of fish from March, June, July and August]{Some samples from the different time periods. We have good conditions during March, but complete darkness at night. During June there is large amounts of algae on the lens. Between June and July the lens was cleaned and the camera angle changed}
    \label{fig:images_with_fish}
\end{figure*}

In \autoref{fig:images_with_fish} we see a set of images containing fish from March to August. It can be seen that the backgrounds vary a lot and that the size and illumination of the fish is not constant. Especially during nighttime we see that some fish appear almost completely white. As we only train our algorithm on data from daytime during March, we expect it to be inaccurate when running detection on images from June, July and August due to the aforementioned reasons. The norm for all \gls{yolo}-libraries is to display training per iterations and not per epochs. However converting from iterations to epochs is quite simple and is shown in \autoref{eq:epochs}.
\begin{equation}
    \text{epochs} = \frac{\text{batchsize} * \text{iterations}}{I_n}
    \label{eq:epochs}
\end{equation}
The batch size is set to 64 and we have $I_n$ images. Precision, recall and F1 are calculated at a lower confidence thresholds of 0.25. Since only one class is being detected \gls{map} and \gls{ap} produces identical values and thus only \gls{map} is displayed.

\paragraph{Training Stage One:}
In \autoref{table:yolo_results} we present the results achieved when training YOLO on our first set of hand labelled images. Based on  \autoref{eq:epochs} we find that 1000 iterations corresponding to roughly 140 epochs.  We see that it only takes about 2000 iterations or so for the algorithm to reach its peak performance. We even see that the algorithm performs worse across all the metrics at 3000 iterations. 

\begin{table}
\centering
    \begin{tabular}{|
    >{\columncolor[HTML]{EFEFEF}}l |l|l|l|l|}
    \hline
    \multicolumn{1}{|c|}{\cellcolor[HTML]{C0C0C0}Iterations} & \multicolumn{1}{c|}{\cellcolor[HTML]{C0C0C0}mAP@50} & \cellcolor[HTML]{C0C0C0}Precision & \cellcolor[HTML]{C0C0C0}Recall & \cellcolor[HTML]{C0C0C0}F1 \\ \hline
    1000   & 0.3881       & 0.53      & 0.50       & 0.51   \\ \hline
    2000      & 0.6003  & 0.72  & 0.64     & 0.68     \\ \hline
    3000 &     0.5974 &  0.72        &    0.61    &       0.66      \\ \hline
    4000    &      0.6015    &   0.73  &  0.62   &   0.67      \\ \hline
    4200      & 0.6338 &0.74 & 0.69 &    0.72       \\ \hline
      
    \end{tabular}
  
    \caption{Metrics from training stage one calculated from the test data consisting of 51 images. All data was hand labeled}
    \label{table:yolo_results}
\end{table}

\begin{table}
    \centering
    \begin{tabular}{|
        >{\columncolor[HTML]{EFEFEF}}l |l|l|l|l|}
        \hline
        \multicolumn{1}{|c|}{\cellcolor[HTML]{C0C0C0}Iterations} & \multicolumn{1}{c|}{\cellcolor[HTML]{C0C0C0}mAP@50} & \cellcolor[HTML]{C0C0C0}Precision & \cellcolor[HTML]{C0C0C0}Recall & \cellcolor[HTML]{C0C0C0}F1 \\ \hline
        1000 & 0.8118 & 0.83 & 0.81 & 0.82\\ \hline
        2000 & 0.8274 & 0.79 & 0.84 & 0.81\\ \hline
        3000 & 0.8595 & 0.83 & 0.88 & 0.85 \\ \hline
        4000 & 0.8679 & 0.82 & 0.87 & 0.85 \\ \hline
        5000 & 0.8475 & 0.85 & 0.84 & 0.84\\ \hline
        - &&&&\\ \hline
        8000 & 0.8809 & 0.83 & 0.87 & 0.85 \\ \hline
    \end{tabular}
    \caption{Metrics from training stage two calculated from the data data consisting of $\sim$ 300 images. All data was pseudo-labeled and then manually corrected}\label{table:yolo_results_more}
\end{table}

2000 iterations correspond to roughly 240 epochs which should be sufficient for a dataset of 500 or so images with only one class to detect. However we ran the algorithm for a little while longer just to be sure that no more improvements could be made. We ended up using the weights learned at iteration 4200 since these scored the best metrics. We see that the algorithm was able to recall 69 \% of the fish in the test images from March with a precision of 74 \% which we believe is quite decent for a dataset of this size. A further strengthener of this belief is the fact that many of these 500 or so images do not contain any fish at all (close to half of the images were just background).

\paragraph{Training Stage Two:}


In \autoref{table:yolo_results_more} we have shown the progress during training stage two, where $\sim 3000$ images were used during the training. Due to the increased dataset size 1000 iterations correspond to roughly 24 epochs. We see from \autoref{table:yolo_results_more} that the increase in size of the dataset greatly improved the metrics across the board. Achieving a \gls{map} of 0.88 and F1-score of 0.85 does not seem too bad given the noisy dataset. It is, however, reasonable to think that better results could be achieved. The limiting quality of the data is probably due to the authors inability to correctly label data by hand. During the manual labeling of the data it happened quite frequently that to decide whether a certain object was a fish or not was infeasible. This led to some ambiguities in the dataset labels. Sometimes a small speck would be labeled as a fish and other times not. Without having any concrete data, from which to conclude from, we informally estimate that the manual labeling was close to 95\% correct and thus postulate that the \gls{map} and F1-scores could be higher. 
We observe that most of the learning is done in the first 1000 iterations. However, we do see a slowly decreasing trend from iteration 1000 to 8000 or so motivating storing the weights from iteration 8000 onwards. Prior to the first stage of training, the part of the network was initialized with weights trained on the ImageNet database. This database does not contain any fish though it contains other $80$ different classes. So in a way what we did in the second stage was transfer learning. Prior to the second stage of training, the weights learned during the first stage were used as initial values. This might be a part of the explanation of why so many more epochs were needed during the first stage of training. We recall that the first stage of training needed about 2000 iterations before giving valuable results while the second stage only needed a bit more than a 1000. 2000 iterations in the first stage correspond to roughly 280 epochs while 2000 iterations in the second stage correspond to roughly 48 epochs. This is significantly less. These results can however, also be explained in part by the fact that the first stage of training was on much lesser data. It is thus tempting to believe that this type of transfer learning accelerated the learning process. We note that using \gls{yolo} to pseudo-label all the images and then manually correcting deviations was a lot faster than manually labelling every image from scratch.

\subsection{Visual Predictions on Unseen Images}
\begin{figure}
    \centering
    \includegraphics[width=1\columnwidth]{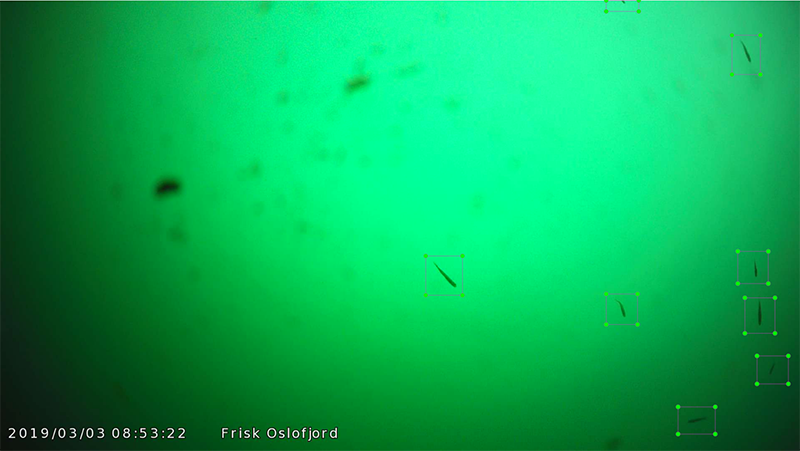} 
    \caption{Image from 03.03.2019 with predicted bounding boxes}
    \label{fig:fish_with_labels}
\end{figure}

In \autoref{fig:fish_with_labels} some detections made on unseen data from March can be seen. It is clear that the algorithm is able to satisfactorily detect the fishes in the image. This is to be expected as an F1-score of $0.85$ was achieved after the second round of training. When inspecting all the data from March we concluded that images from throughout the month were similar and hence the training set was indeed representative of the test set. This is reflected in excellent performance of the algorithm on unseen data from March (like in \autoref{fig:fish_with_labels})


\begin{figure}
    \centering
        \begin{subfigure}{1\columnwidth}
        \centering
        \includegraphics[width=1\columnwidth]{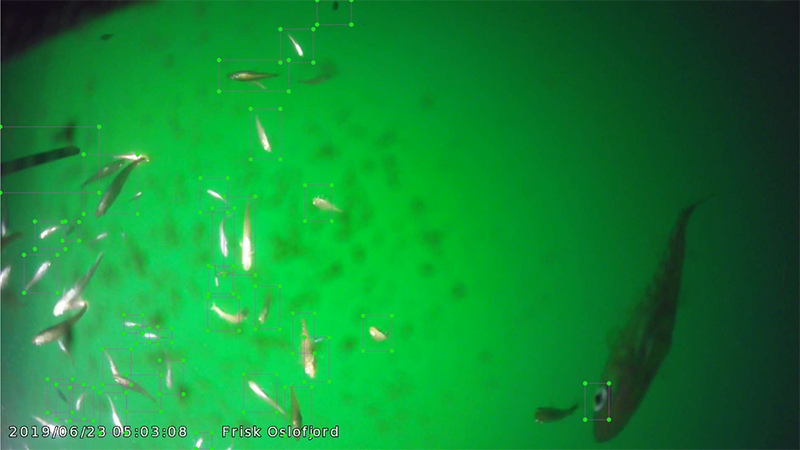}
        \caption{}
        \label{det_a}
    \end{subfigure}
    \begin{subfigure}{1\columnwidth}
        \centering
        \includegraphics[width=1\columnwidth]{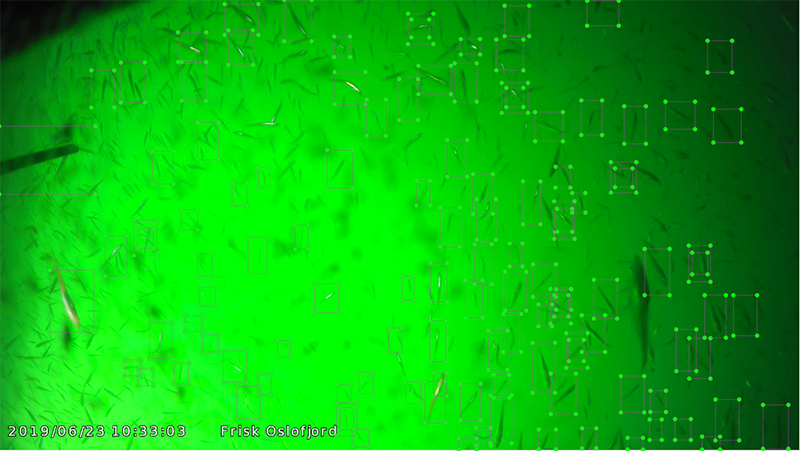}
        \caption{}
        \label{det_b}
    \end{subfigure}
    \begin{subfigure}{1\columnwidth}
        \centering
        \includegraphics[width=1\columnwidth]{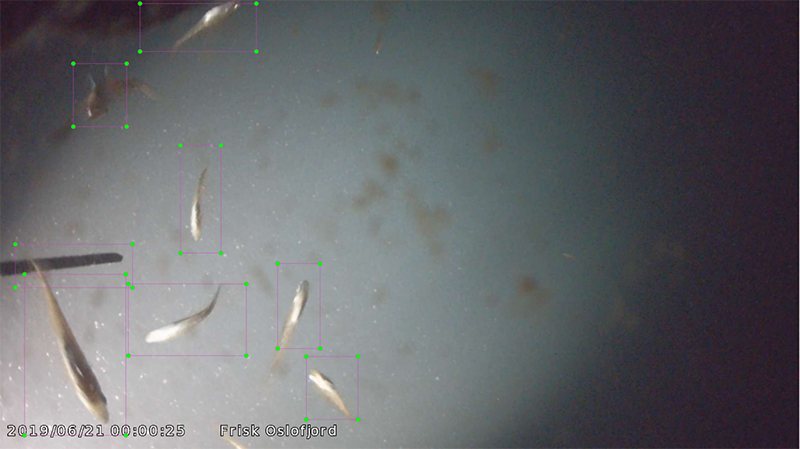}
        \caption{}
        \label{det_c}
    \end{subfigure}
    \caption{Some predictions on data sampled from 23. June 2019}
    \label{fig:fish1_detections}
\end{figure}

When the model trained on data from March was applied to data from other months (eg. June and July), mixed results were obtained as can be seen in \autoref{fig:fish1_detections}. On the left hand side of Figures \ref{det_a}, \ref{det_b} and \ref{det_c} it looks like a twig or perhaps a piece of plastic from the rig is being marked as a fish. This indicates that the trained model is not sufficiently robust to outliers and noise, and has not learned well enough what constitutes a fish. However, this does make sense as no noise like this was seen in the training dataset from March. Another example of this is in \autoref{det_a} where the eye of a large fish is marked as a fish. Furthermore, in some of the images in \autoref{fig:fish1_detections} we see a large number of fishes got successfully detected. Especially in \autoref{det_b} we notice that the trained model did fairly well in detecting every tiny fishes in abundance. However, at the same time there are also a large number of fishes that go undetected. We remark that the precision is very high while recall is quite low. Almost all of the bounding boxes that are placed are correctly placed around fishes. We wondered if the reason that not more fish was detected was because of a limited amount of bounding boxes. However, at the largest scale in the network $52 \times 52 \times 3$ bounding boxes are generated and that should be sufficient. The two potential reasons why the model behaves poorly on certain dataset can be:
\begin{enumerate}
    \item The images in \autoref{fig:fish1_detections} are overall quite different from the data the model was trained on. Thus, all the different environmental conditions unseen during the training step disturb the models ability to correctly detect fishes.
    \item  When images are input to \gls{yolo} they are downscaled to $416 \times 416$. This results in loss of resolution and hence loss of information required for correct detection and classification of the fishes.
\end{enumerate}{}
 
\begin{figure}
    \centering
    \begin{subfigure}{1\columnwidth}
        \centering
        \includegraphics[width=1\columnwidth]{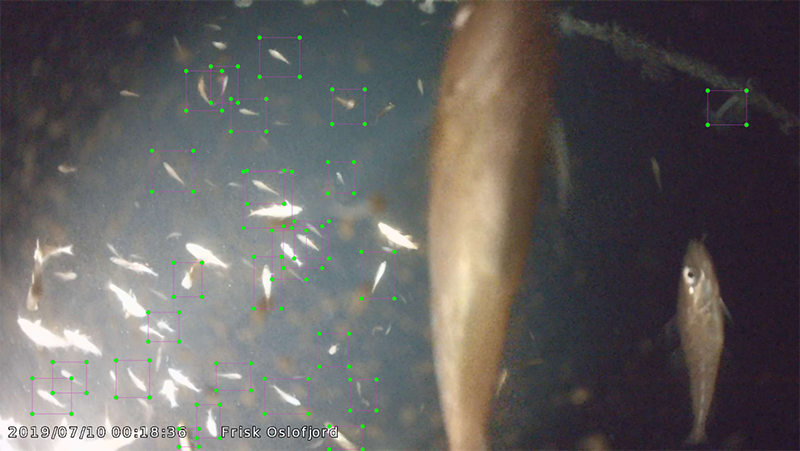}
        \caption{Nighttime 10.07.2019}
        \label{dete_a}
    \end{subfigure}
    \begin{subfigure}{1\columnwidth}
        \centering
        \includegraphics[width=1\columnwidth]{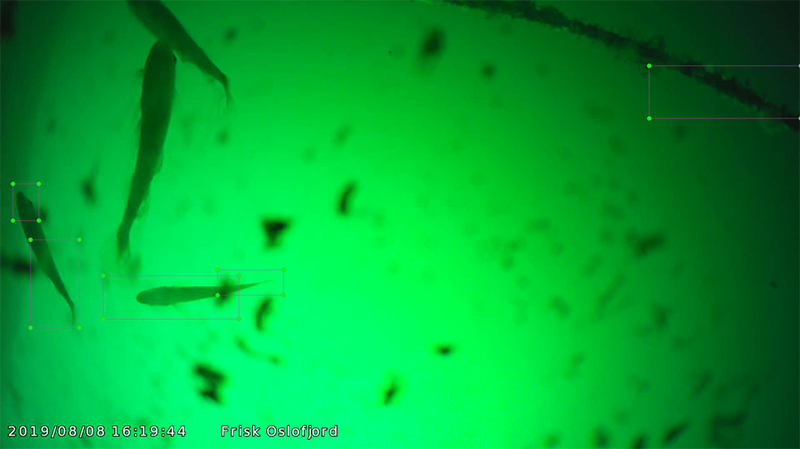}
        \caption{Daytime 08.08.2019}
        \label{dete_b}
    \end{subfigure}
    \caption{Some especially poor detections from July and August}
    \label{fig:fish_bad_det}
\end{figure}

In \autoref{fig:fish_bad_det} we present some examples where our algorithm performed poorly. It is not able to recognize large fish since there were no large fish in its training data and it can not generalize enough from the data it was given. In addition to the absence of a few detections it also occasionally gives two bounding boxes to one fish as seen in \autoref{dete_b}. Furthermore, the rope seen in this image is being detected as a fish. From this we know that the algorithm does not have to see an entire fish to mark it as a fish. This is good in the sense that it can detect fish that are not completely within the field of view of the camera but leads to several errors as shown here. If data containing large fish was included in the training set this could probably have been remedied to some degree. If a large fish would be detected, \gls{nms} would suppress the smaller bounding boxes only encapsulating part of the fish. The algorithm never marks algae as fish which is impressive. For example in \autoref{dete_b} it could have been a possibility to mistake algae for fish. Why the algorithm does not do this is unclear. Perhaps it uses the blurriness of the algae to determine they are not fish. Or perhaps it is all in the shape. It would be interesting to see how the algorithm retrained on more classes behave. Last but not least we note that the algorithm is able to detect fish during a variety of lighting conditions as seen in \autoref{fig:fish1_detections}. We never trained our algorithm on data that had artificial lighting, but still the algorithm detects most of the fish that are illuminated with artificial light. 

\subsection{Insight into the inner workings of the FCN}\label{subsec:insight_pca}
\begin{figure*}
    \centering
    \begin{subfigure}{0.5\textwidth}
        \centering
        \includegraphics[width=1\columnwidth]{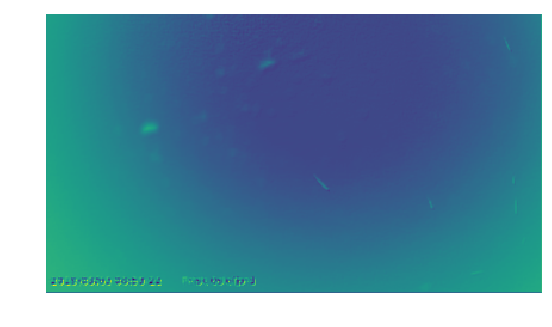}
        \caption{}
        \label{first2_a}
    \end{subfigure}%
    \centering
    \begin{subfigure}{0.5\textwidth}
        \centering
        \includegraphics[width=1\columnwidth]{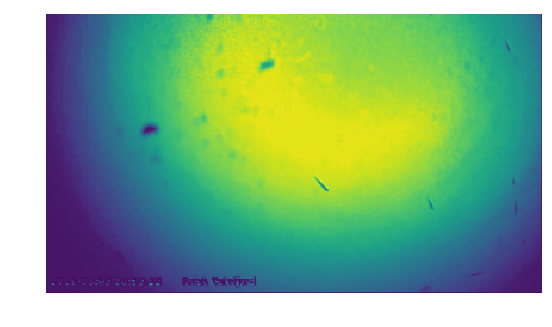}
        \caption{}
        \label{first2_b}
    \end{subfigure}
    \begin{subfigure}{0.5\textwidth}
        \centering
        \includegraphics[width=1\columnwidth]{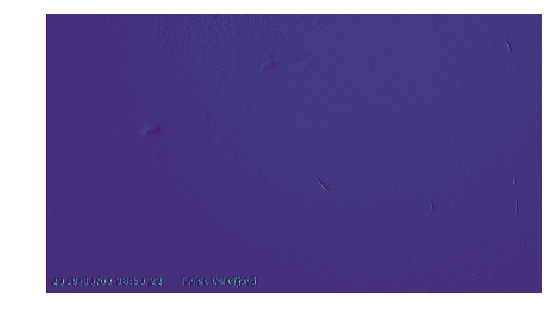}
        \caption{}
        \label{first2_d}
    \end{subfigure}%
   \begin{subfigure}{0.5\textwidth}
        \centering
        \includegraphics[width=1\columnwidth]{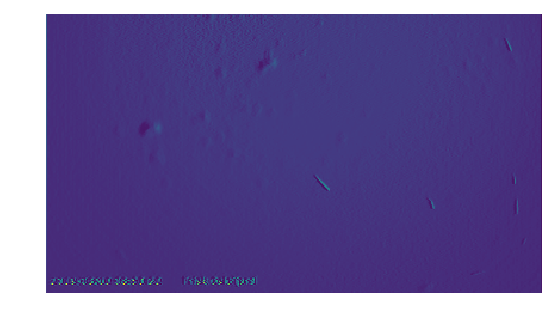}
         \caption{}       
        \label{first2_e} 
    \end{subfigure}
    \begin{subfigure}{0.5\textwidth}
        \centering
        \includegraphics[width=1\columnwidth]{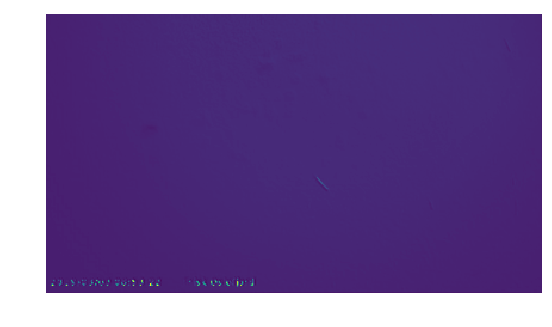}
        \caption{}
        \label{first2_f}
    \end{subfigure}%
    \begin{subfigure}{.5\textwidth}
        \centering
        \includegraphics[width=1\textwidth]{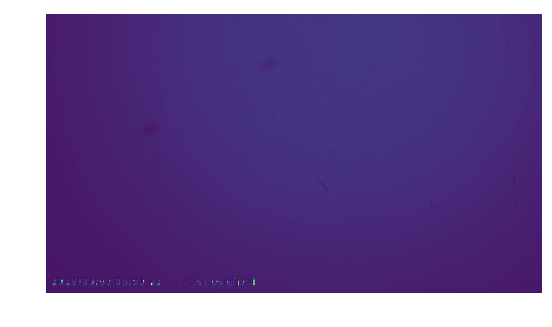}
         \caption{}       
        \label{first2_g} 
    \end{subfigure}
    \caption{Images extracted from the first convolutional layer in the network}
    \label{fig:layer1_results}
\end{figure*}

In order to understand the inner workings of the network, feature maps from the hidden layers were extracted. From the description of the feature extractor in
\cite{redmon2018yolov3} it is known that the first layer of the network has $32$ filters of size $3 \times 3$. In reality these filters are $3 \times 3 \times 3$ because color images have three channels. It should be noted that $32$ is actually the lowest number of filters in any layer. Some layers in the network  has up to $1024$ filters making the task of visualizing and interpreting them individually almost impossible. It is worth stressing that the result produced by the convolution operation on the colored images do not actually produce images that can be visualized in a comprehensible way. This is because the resulting matrix values are not confined to the 0-255 interval.  Therefore, to actually create visualisations, the values were normalized and the default colormap ``viridis'' from Matplotlib was applied. The colormap maps low values to dark blue and high values to yellow. The image that was fed as input to the trained network from which intermediate feature maps were extracted is given by \autoref{fig:fish_with_labels}. In \autoref{fig:layer1_results} we see some plots of the intermediate images produced in the very first convolutional layer of the network. Based on these images it seems that the filters produce every imaginable variant of the image. Some filters blur the image while others sharpen it. Some even seems to produce the negative. In \autoref{first2_e} there are strong gradients highlighting edges while in \autoref{first2_f} the image is almost completely smooth. In Figures \ref{first2_d} and \ref{first2_e} one can see that edges are detected on opposite sides. In \autoref{first2_d} left-edges are detected, while in \autoref{first2_e} right-edges are. In Figures \ref{first2_a} and \ref{first2_b} we see inverse values. Using the negative might be one of the reasons why the network seems to detect fish both with and without the presence of artificial lighting. This also explains why it doesn't matter to the network whether the fishes are dark or light in color. 

\begin{figure*}
    \centering
    \begin{subfigure}{.25\textwidth}
        \centering
        \includegraphics[width=1\textwidth]{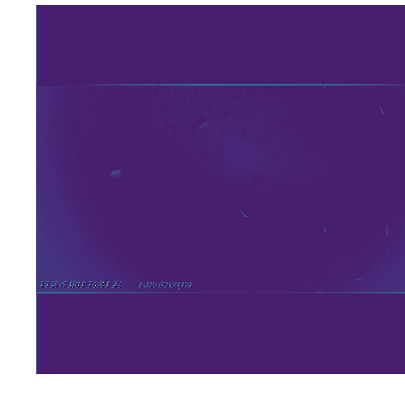}
        \caption{L1: conv}
        \label{fig:int_a}
    \end{subfigure}%
    \begin{subfigure}{.25\textwidth}
        \centering
        \includegraphics[width=1\textwidth]{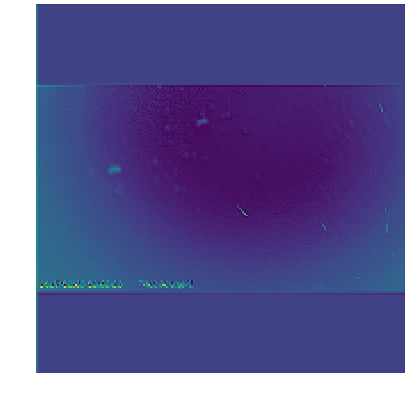}
        \caption{L2: conv}
        \label{fig:int_b}
    \end{subfigure}%
    \begin{subfigure}{.25\textwidth}
        \centering
        \includegraphics[width=1\textwidth]{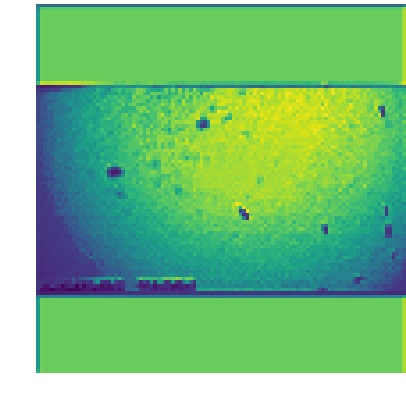}
        \caption{L6: conv}
        \label{fig:int_c}
    \end{subfigure}%
    \begin{subfigure}{.25\textwidth}
        \centering
        \includegraphics[width=1\textwidth]{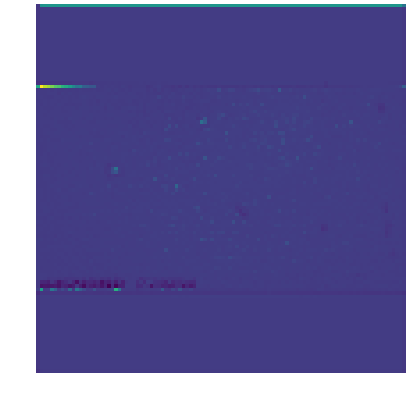}
        \caption{L7: conv}
        \label{fig:int_d}
    \end{subfigure}
    
    \begin{subfigure}{.25\textwidth}
        \centering
        \includegraphics[width=1\textwidth]{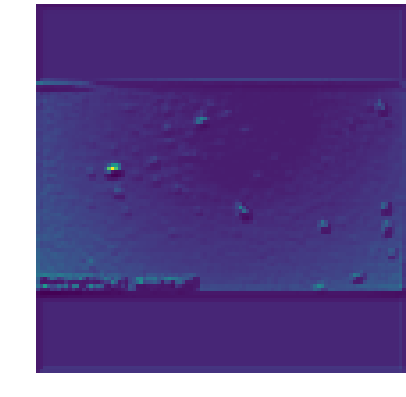}
         \caption{L8: conv}       
        \label{fig:int_e}
    \end{subfigure}%
    \begin{subfigure}{.25\textwidth}
        \centering
        \includegraphics[width=1\textwidth]{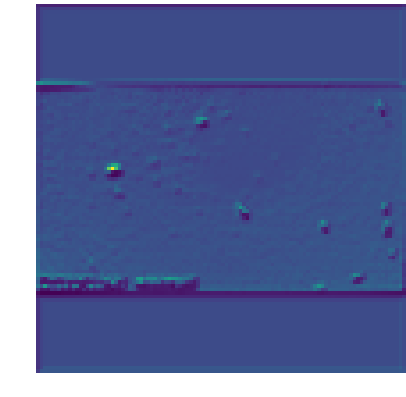}
        \caption{L9: shortcut}
        \label{fig:int_f}
    \end{subfigure}%
    \begin{subfigure}{.25\textwidth}
        \centering
        \includegraphics[width=1\textwidth]{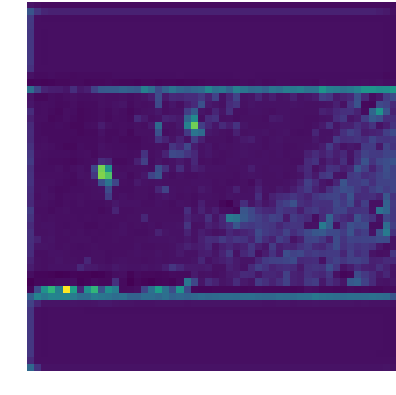}
         \caption{L20: conv}       
        \label{fig:int_g} 
    \end{subfigure}%
    \begin{subfigure}{.25\textwidth}
        \centering
        \includegraphics[width=1\textwidth]{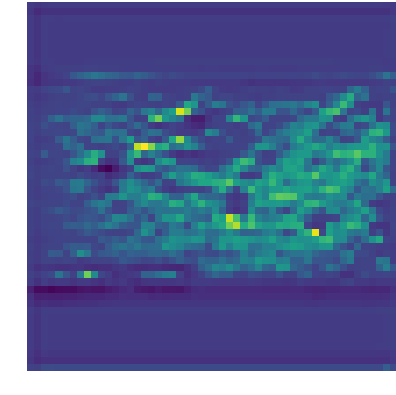}
        \caption{L21: conv}
        \label{fig:int_h}
    \end{subfigure}

    \begin{subfigure}{.25\textwidth}
        \centering
        \includegraphics[width=1\textwidth]{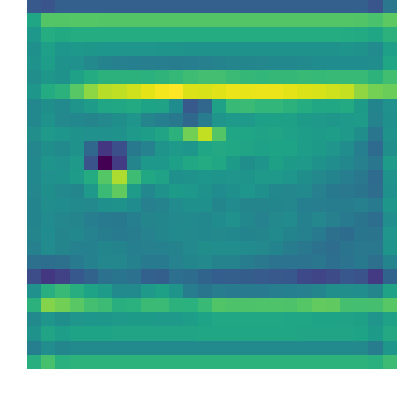}
         \caption{L40: conv}       
        \label{fig:int_i} 
    \end{subfigure}%
    \begin{subfigure}{.25\textwidth}
        \centering
        \includegraphics[width=1\textwidth]{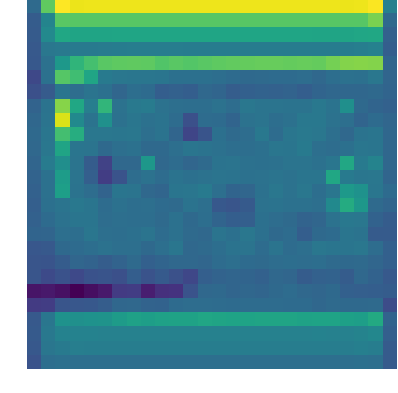}
        \caption{L41: shortcut}
        \label{fig:int_j}
    \end{subfigure}%
    \begin{subfigure}{.25\textwidth}
        \centering
        \includegraphics[width=1\textwidth]{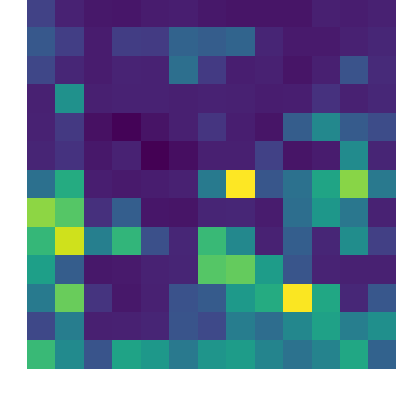}
        \caption{L81: conv}
        \label{fig:int_l}
    \end{subfigure}%
     \begin{subfigure}{.25\textwidth}
        \centering
        \includegraphics[width=1\textwidth]{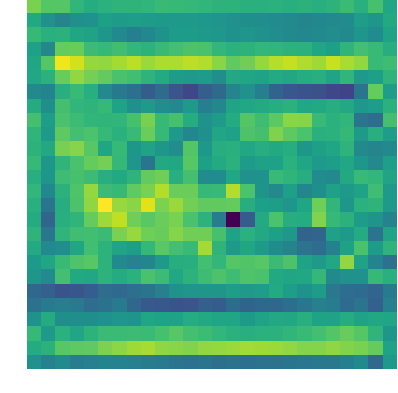}
        \caption{L94: conv}
        \label{fig:int_90n}
    \end{subfigure}
    
    \begin{subfigure}{.25\textwidth}
        \centering
        \includegraphics[width=1\textwidth]{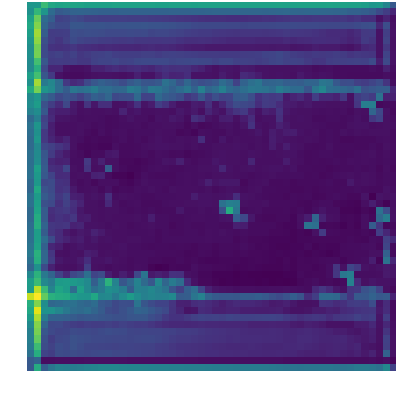}
         \caption{L103: conv}       
        \label{fig:int_101o} 
    \end{subfigure}%
    \begin{subfigure}{.25\textwidth}
        \centering
        \includegraphics[width=1\textwidth]{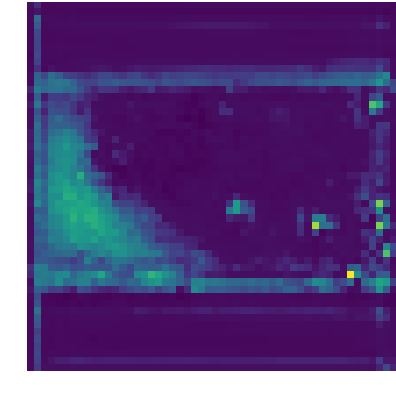}
        \caption{L104: conv}
        \label{fig:int_102p}
    \end{subfigure}%
      \begin{subfigure}{.25\textwidth}
        \centering
        \includegraphics[width=1\textwidth]{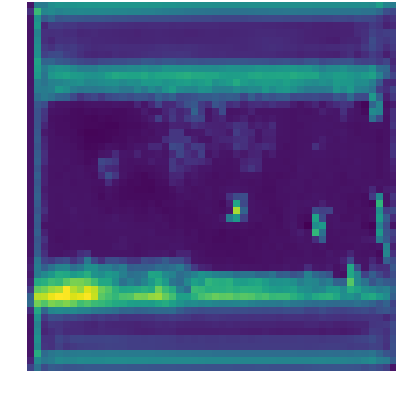}
        \caption{L105: conv}
        \label{fig:int_103q}
    \end{subfigure}%
    \begin{subfigure}{.25\textwidth}
        \centering
        \includegraphics[width=1\textwidth]{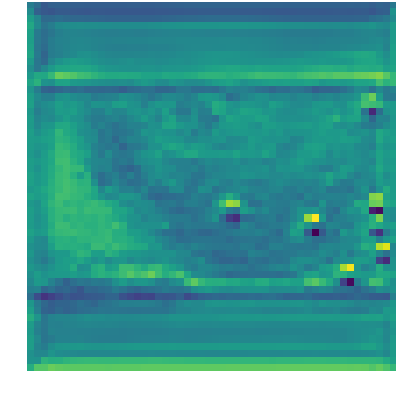}
        \caption{L106: conv}
        \label{fig:int_104r}
    \end{subfigure}
    \caption[Feature maps from intermediate layers in YOLO]{Feature maps from intermediate layers in YOLO. Detections are made at layer 82, 94 and 106}
    \label{fig:traveling_through_yolo}
\end{figure*}

In Figure \ref{fig:traveling_through_yolo} images reconstructed in deeper layers are displayed. When comparing Figures \ref{fig:layer1_results} and \ref{fig:traveling_through_yolo} one notices that the images have different aspect ratios. When looking at the results from the first layer one can look at the results using the original aspect ratio of the images that were put in. However, as an image passes through the network, more and more information seems to ``bleed'' onto the originally unused top and bottom margins of the image. It seems like YOLO tries to store as much information as possible. 

In the layers closer to the input layer (e.g. Figures \ref{fig:int_a}, \ref{fig:int_b}, \ref{fig:int_c}) one can see that the images closely resemble the original image. Gradually as one traverses through the layers the images become coarser. In Figures \ref{fig:int_e} and \ref{fig:int_f} one can see that the fish and some algae are very prominent. In these images the algae is brighter than the fish. It might be that high values in these specific images in these specific layers means that an object should be ignored, as we do not want labels around algae. In Figure \ref{fig:int_l} a reconstructed image from the $81^{st}$ layer of the network is shown. This layer consists of $1024$ feature maps of size $13 \times 13$. It is based on the images in this layer that the first prediction is made. When we have traversed this far into the network it is virtually impossible to recognize what information the different pixels encode. However, it is to be noted that the original aspect ratio border seems to have vanished, and that information seems to be stored in the entire image. One can recall that YOLO makes detections at three different scales. As one moves past the first detection layer the network starts to scale up the image. In Figure \ref{fig:int_90n} one can see an image from the layer before the second detection. It is still virtually impossible for a human to harness any meaningful information from these images. However, one sees that some horizontal lines have started to appear. It seems like YOLO is starting to reconstruct the original image. After the second detection layer YOLO further scales up the image. In the four layers prior to the third and last detection, the images start to make a bit more sense. In Figure \ref{fig:int_101o} one can recognize blobs that correspond to the fish that is to be detected. It is evident that YOLO is able to ``remember'' what the original image looked like. From this layer onwards, and to the end, these blobs become more and more distinct as can be seen in Figures \ref{fig:int_102p}, \ref{fig:int_103q} and \ref{fig:int_104r}. Thus for the very last detection layer the network can perhaps create bounding boxes around the brightest pixels. Furthermore, one can notice that in Figure \ref{fig:int_104r} the fishes are represented by blocks of bright and dark pixels together. It can also be observed that the restored top and bottom margins of these images contain very little variance. It seems that in this detection layer most of the information is retrieved from the values within the original aspect ratio.

\begin{figure*}
    \centering
    \begin{subfigure}{.3\textwidth}
        \centering
        \includegraphics[width=1\textwidth]{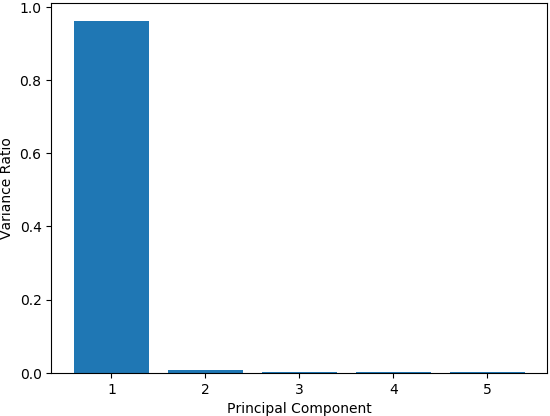}
         \caption{L1: N=32}       
        \label{fig:pca_1} 
    \end{subfigure}%
    \begin{subfigure}{.3\textwidth}
        \centering
        \includegraphics[width=1\textwidth]{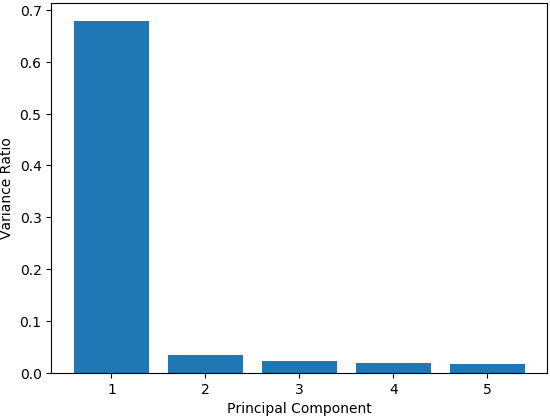}
        \caption{L2: N=64}
        \label{fig:pca_2}
    \end{subfigure}%
    \begin{subfigure}{.3\textwidth}
        \centering
        \includegraphics[width=1\textwidth]{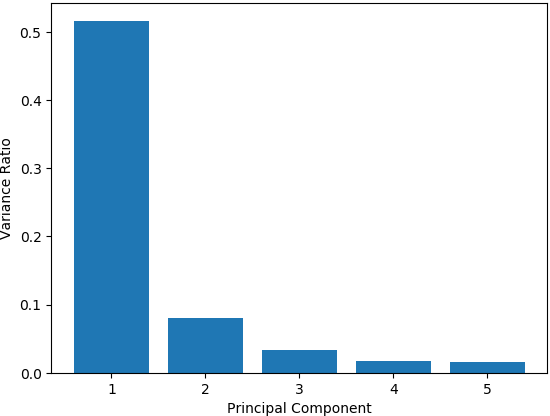}
         \caption{L10: N=64}       
        \label{fig:fig:pca_10} 
    \end{subfigure}
    \begin{subfigure}{.3\textwidth}
        \centering
        \includegraphics[width=1\textwidth]{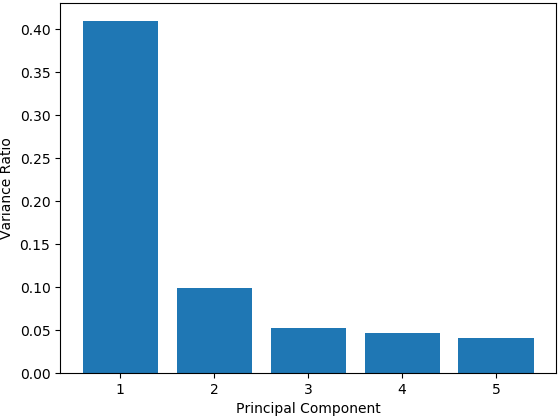}
        \caption{L20: N=128}
        \label{fig:fig:pca_11}
    \end{subfigure}%
      \begin{subfigure}{.3\textwidth}
        \centering
        \includegraphics[width=1\textwidth]{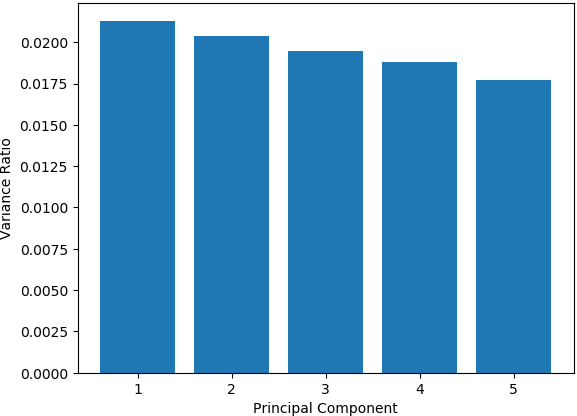}
        \caption{L81: N=1024}
        \label{fig:fig:pca_81}
    \end{subfigure}%
    \begin{subfigure}{.3\textwidth}
        \centering
        \includegraphics[width=1\textwidth]{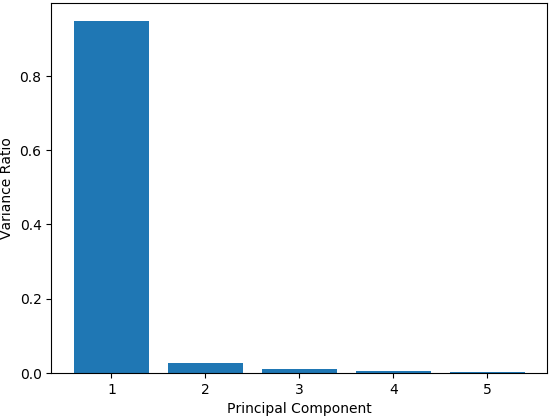}
        \caption{L82: N=18}
        \label{fig:fig:pca_82}
    \end{subfigure}
    
    \caption[Bar plots: PCA analysis of intermediate layers]{Bar plots: PCA analysis of intermediate layers. N is the number of feature maps.}
    \label{fig:travel_pca}
\end{figure*}

The discussion so far is based on a handful of images extracted from each layer. As explained earlier there can be up to 1024 feature maps in some of the layers which are humanly impossible to interpret. In order to develop some statistical understanding of different layers we conducted \gls{pca} on the images extracted from each layers individually \cite{pca}. \autoref{fig:travel_pca} gives the plots of the ratio of variance for the 5 most prominent principal components. It appears that for our early layers, as can be seen in Figures \ref{fig:pca_1} and \ref{fig:pca_2}, almost all the information can be explained using a single component. Gradually, as we move through the layers it seems that more and more information is spread out across the images within a layer. In other words the feature maps become more and more distinct within each layer. All the images in the first layer are quite alike, while the images are all very different in the $81^{st}$ layer. However, in the next layer (Figure \ref{fig:fig:pca_82}), a mode collapse if observed. Seemingly, the output filters are able to extract some pattern from the $81^{st}$ layer. If all the components were equally contributing that could indicate that no pattern was found in the data. Perhaps the filters in the early layers perform similar operations, while in later layers the operations become more specialised. If this is the case it would make sense that the deeper layers contain more distinct data. We recall that YOLO is a large FCN capable of detecting and classifying several thousand classes simultaneously and that it might be overkill to just detect one class, as is being doing here. Perhaps PCA would give very different results if the network was trained on a different dataset with more classes. It might be that this kind of PCA could be used as an optimization technique on FCNs. If most of the variance is explained by one, or a few, principal components, perhaps the number of filters in that layer could be reduced. This could be implemented to reduce run times of FCNs. This could be done by first training the network. Then running PCA and reducing the amount of filters in the layers that are mostly explained by a few principal components. Then the network could be retrained and PCA re-calculated. This could be done until the accuracy starts to drop below a certain threshold, in relation to the original accuracy of the network. This would greatly increase training times, but could speed up test times while still maintaining almost the same accuracy as the original network. This could in addition make it easier to interpret and explain the network as a simplified network is nevertheless easier to analyze.

\begin{figure*}
    \centering
    \begin{subfigure}{.3\textwidth}
        \centering
        \includegraphics[width=1\textwidth]{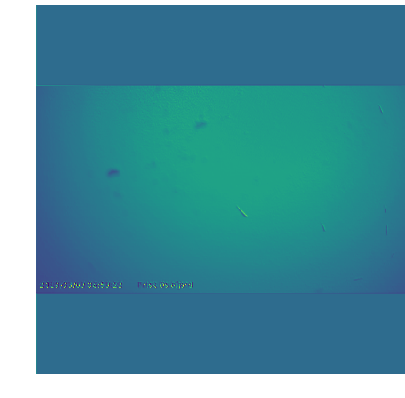}
         \caption{L1: First eigenvector}       
        \label{fig:pcai_11} 
    \end{subfigure}%
    \begin{subfigure}{.3\textwidth}
        \centering
        \includegraphics[width=1\textwidth]{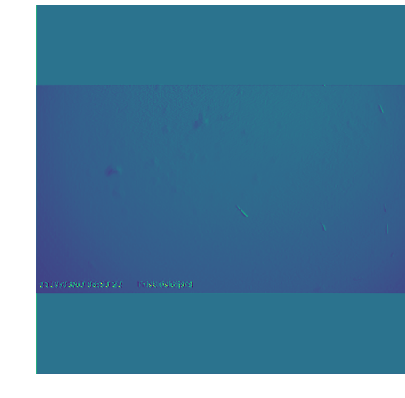}
        \caption{L1: Second eigenvector}
        \label{fig:pcai_12}
    \end{subfigure}%
    \begin{subfigure}{.3\textwidth}
        \centering
        \includegraphics[width=1\textwidth]{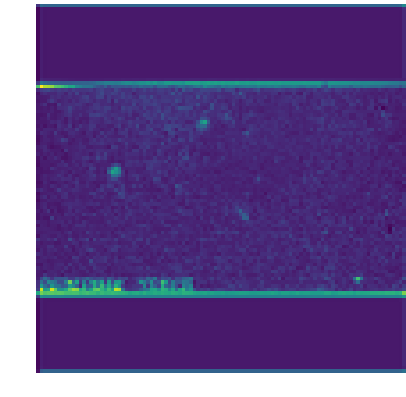}
         \caption{L10: First eigenvector}       
        \label{fig:fig:pcai_101} 
    \end{subfigure}
    \begin{subfigure}{.3\textwidth}
        \centering
        \includegraphics[width=1\textwidth]{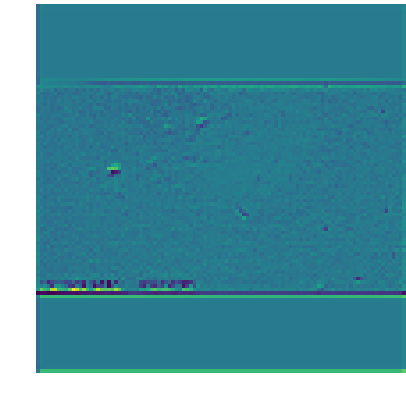}
        \caption{L10: Second eigenvector}
        \label{fig:fig:pcai_102}
    \end{subfigure}%
    \begin{subfigure}{.3\textwidth}
        \centering
        \includegraphics[width=1\textwidth]{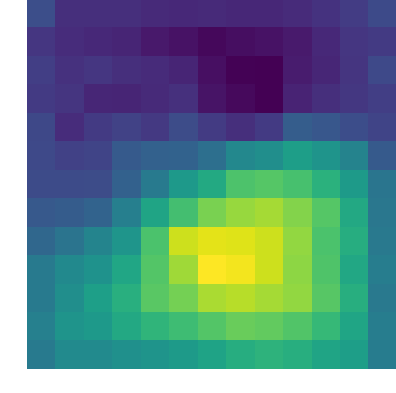}
        \caption{L81: First eigenvector}
        \label{fig:fig:pcai_811}
    \end{subfigure}%
    \begin{subfigure}{.3\textwidth}
        \centering
        \includegraphics[width=1\textwidth]{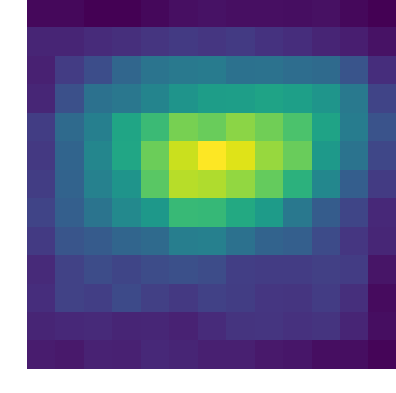}
        \caption{L81: Second eigenvector}
        \label{fig:fig:pcai_812}
    \end{subfigure}
    \caption[Images: PCA analysis of intermediate layers]{Images: PCA analysis of intermediate layers}
    \label{fig:travel_pcai_imgs}
\end{figure*}

In \autoref{fig:travel_pcai_imgs} we see visualizations made by constructing images from only the most, and second most, important components produced by the PCA. In Figure \ref{fig:pca_1} we observe that by just using the first component we retain almost all the variance. This is in correspondence with what we see in Figures \ref{fig:pcai_11} and \ref{fig:pcai_12}. There is seemingly very little information in the second image that we can not be found in the first. We see from Figures \ref{fig:fig:pcai_101} and \ref{fig:fig:pcai_102} that the images have started to become more distinct. In Figure \ref{fig:fig:pca_10} we see that about 50\% of the variance in this image is explained by the first principal component. 
In the $81^{st}$ layer, as seen in Figures \ref{fig:fig:pcai_811} and \ref{fig:fig:pcai_812}, we see that the principal components starts to hone in on certain regions. This is especially interesting when we look at a normal image from the same layer, as seen in \autoref{fig:int_l}, which mostly looks like noise. In this layer the principal components explain almost the same amount of variance as can be seen in \autoref{fig:fig:pca_82}. 

\section{Conclusion}
In this paper we utilized YOLO to detect fish in images recorded under water and provided insight into the internal workings of the algorithm. The major findings of the project can be enumerated as follows:
\begin{itemize}
    \item The work presented in this report makes a significant contribution to the Healthy Oslo fjord project by building a workflow that can be used to generate labelled data from images using semi-supervised learning. The current trained model already has a F1-score of 0.85 and has been used to generate labelled dataset that will also provide labels to the sonar data that was acquired during the same field campaign.
    \item To use unsupervised learning when pseudo-labeling used to be the popular method \cite{sermanet2012pedestrian}. We showed that using \gls{yolo} as a semi-supervised algorithm bears some merit. Using a tiny amount of hand labeled data to pseudo-label a lot of data greatly reduced the manual labor involved. 
    \item The project led to a better insight into the inner working of the YOLO algorithm. Internal layers of YOLO network were illustrated and multivariate data analysis tools like PCA was utilized to extract information from the thousands of filters used in the network. Investigating the variance ratio vs principle component plots for the different layers, we observed that most of the variance in the reconstruction of the images using the trained filters in the first few layers can be explained by a single component. The number of components required to explain the variance in the deep layers gradually increases. 
\end{itemize}

Despite the interesting results presented in the article, there are several aspects in the current work that requires more in-depth investigation. We used a very complicated network to detect only a single class (of fish). It would be interesting to see how the network behaves if it is utilized to do multiclass object detection and classification. Also it will be interested to explore the possibility of employing PCA to optimize the network. 

\section*{Acknowledgement}
The authors would like to thank the Healthy Oslo Fjord initiative for providing the funds that enabled creating the dataset which we have presented here.

\bibliographystyle{cas-model2-names}
\bibliography{cas-refs}

\newpage

\bio{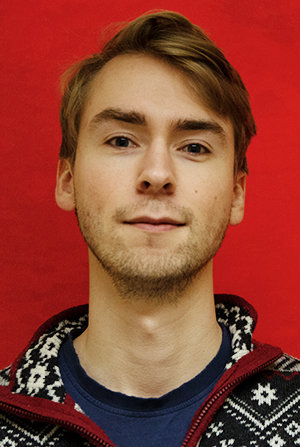}
Herman Stavelin is a Masters student in Robotics and Cybernetics at the Norwegian University of Science and Technology. He is currently writing a Masters Thesis on Object Detection and Explainable AI. 
\endbio
\vskip50pt
\bio{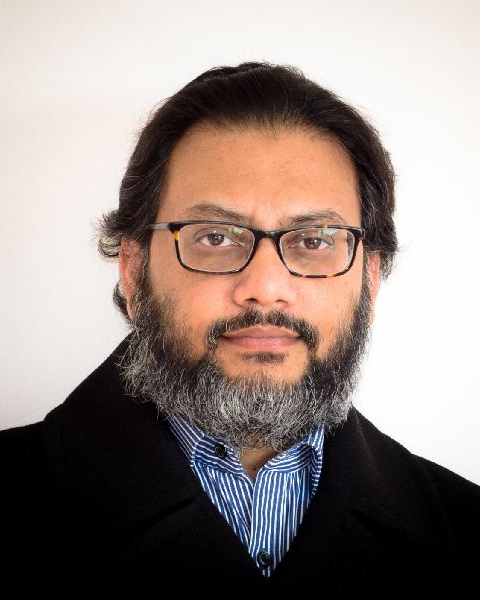}
Adil Rasheed is the professor of Big Data Cybernetics in the Department of Engineering Cybernetics at the Norwegian University of Science and Technology where he is working to develop novel hybrid methods at the intersection of big data, physics driven modelling and data driven modelling in the context of real time automation and control. He also holds a part time senior scientist position in the Department of Mathematics and Cybernetics at SINTEF Digital where he led the Computational Sciences and Engineering group between 2012-2018. He holds a PhD in Multiscale Modeling of Urban Climate from the Swiss Federal Institute of Technology Lausanne. Prior to that he received his bachelors in Mechanical Engineering and a masters in Thermal and Fluids Engineering from the Indian Institute of Technology Bombay.
\endbio
\bio{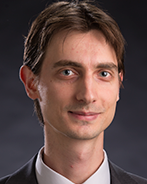}
Omer San received his bachelors in aeronautical engineering from Istanbul Technical University in 2005, his masters in aerospace engineering from Old Dominion University in 2007, and his Ph.D. in engineering mechanics from Virginia Tech in 2012. He worked as a postdoc at Virginia Tech from 2012-'14, and then from 2014-'15 at the University of Notre Dame, Indiana.  
He has been an assistant professor of mechanical and aerospace engineering at Oklahoma State University, Stillwater, OK, USA, since 2015. He is a recipient of U.S. Department of Energy 2018 Early Career Research Program Award in Applied Mathematics. His field of study is centered upon the development, analysis and application of advanced computational methods in science and engineering with a particular emphasis on fluid dynamics across a variety of spatial and temporal scales. 
\endbio
\bio{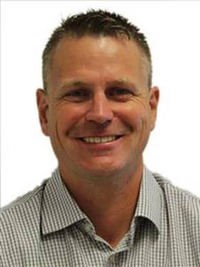}
Arne Hestnes is a software architect at the Kongsberg Maritime Sensor and Robotics division. He is a Master of Technology in machine learning from the Norwegian University of Science and Technology. Currently the main work evolves around connecting industrial sensors to cloud and automating processing flows.
\endbio
\end{document}